\documentclass{article} 
\usepackage{iclr2027_conference,times}

\usepackage{amsmath,amsfonts,bm}

\def\eqref#1{equation~\ref{#1}}

\def\1{\bm{1}}

\DeclareMathAlphabet{\mathsfit}{\encodingdefault}{\sfdefault}{m}{sl}
\SetMathAlphabet{\mathsfit}{bold}{\encodingdefault}{\sfdefault}{bx}{n}

\usepackage[hidelinks]{hyperref}
\usepackage{url}
\usepackage{amssymb,amsthm,mathtools}
\usepackage{booktabs}
\usepackage{multirow}
\usepackage{graphicx}
\usepackage{subcaption}
\usepackage{algorithm}
\usepackage{algorithmic}
\usepackage{tikz}
\usetikzlibrary{arrows.meta,positioning,fit,backgrounds,calc}
\usepackage[capitalize,noabbrev]{cleveref}

\usepackage[disable,textsize=tiny,colorinlistoftodos]{todonotes}

\newcommand{\crfg}{$\mathrm{FedCRF}_{G}$}
\newcommand{\crfv}{$\mathrm{FedCRF}_{F}$}
\newcommand{\crfh}{$\mathrm{FedCRF}_{G+F}$}

\newcommand{\Dval}{\mathcal{D}_{\mathrm{prox}}}

\title{Federated Learning with Energy-Based \\ Structured Probabilistic Inference}

\author{
Daniil Kirilenko\textsuperscript{1,3}\thanks{Equal contribution.} 
\quad
Dario Fenoglio\textsuperscript{1,2}\footnotemark[1] 
\quad
Martin Gjoreski\textsuperscript{1}
\quad
Marc Langheinrich\textsuperscript{1}
\\[0.6em]
\textsuperscript{1}Faculty of Informatics, Universit\`a della Svizzera italiana (USI), Lugano, Switzerland\\
\textsuperscript{2}Massachusetts Institute of Technology (MIT), Cambridge, MA, USA\\
\textsuperscript{3}Northeastern University, Boston, MA, USA
}

\iclrfinalcopy

\begin{document}

\maketitle

\lhead{Preprint}

\vspace{-5pt}
\begin{abstract}
Federated Learning (FL) enables multiple clients to collaboratively train machine learning models without sharing their local data. However, heterogeneity in client data and the presence of unreliable or adversarial updates make effective server aggregation challenging, motivating methods that adapt client contributions according to their reliability and relationships. In this work, we formulate server aggregation in FL as structured probabilistic inference using conditional random fields. We represent clients as nodes in a graph and define unary and pairwise energy potentials. Unary potentials capture individual client properties, while pairwise potentials encode similarities between clients. 
This formulation treats individual client evidence and pairwise relationships as coupled evidence in a joint inference problem, allowing client reliabilities to be inferred collectively and the resulting estimates to guide aggregation.
Where most FL methods target a single challenge, FedCRF's inferred reliabilities and affinities provide one common representation for global aggregation, Byzantine robustness, clustered FL, and personalized FL. With a single set of hyperparameters, it is competitive with or outperforms specialized methods on CIFAR-10, FEMNIST, and Fed-ISIC2019 across diverse forms of client heterogeneity and unreliability.
\end{abstract}

\section{Introduction}
\label{sec:intro}

Federated learning (FL) enables multiple clients to collaboratively train machine learning models under the coordination of a central server while keeping their data local \citep{mcmahan2017communication,kairouz2021advances}. A standard approach is Federated Averaging (FedAvg), 
where clients train locally from a shared model and the server aggregates the resulting models in proportion to their number of training examples \citep{mcmahan2017communication}.
This procedure provides a simple foundation for learning from decentralized data, but sample counts alone do not determine how  
client contributions should be combined.
Differences in data distributions, data quality, and local optimization can produce conflicting updates, while malicious clients may deliberately submit harmful ones. 
The same observed disagreement can therefore arise from different causes and call for different aggregation strategies. 
Consequently, effective aggregation requires the server to account for both the reliability of individual contributions and the relationships between them.


Existing FL methods largely address these cases through specialized mechanisms. 
Client-side approaches constrain local optimization through proximal terms or control variates \citep{li2020federated,karimireddy2020scaffold}, while server-side aggregation methods normalize unequal local work \citep{wang2020fednova} or adapt client weights 
\citep{li2023fedlaw,shi2025fedlws,wang2024fedcda}. Robust aggregation methods limit 
the influence of malicious updates using robust statistics or update selection \citep{blanchard2017machine,yin2018byzantine,pillutla2022robust}. When a single global model cannot adequately represent all clients, clustered FL learns models for groups of related clients \citep{sattler2019cfl,ghosh2020ifca,ruan2022fedsoft,guo2024fedrc}, and personalized FL adapts models to individual clients \citep{zhang2021fedfomo}. 
These approaches typically introduce separate mechanisms for aggregation, robustness, clustering, and personalization, although each relies on assessing which client contributions should be trusted or combined.
Across these settings, existing methods rely on individual client scores or inter-client relationships as direct signals, but do not explicitly model how the two forms of evidence jointly determine client reliability.


In this work, we formulate the server's aggregation decision as \emph{structured probabilistic inference} using a Conditional Random Field (CRF) \citep{lafferty2001conditional}, which we call FedCRF. We represent the participating clients as nodes in a fully connected graph and associate each client with a latent reliability label controlling whether its contribution should be retained, down-weighted, or suppressed. We define a client CRF energy consisting of \emph{unary} and \emph{pairwise} potentials. Unary potentials encode evidence about individual clients, while pairwise potentials encode affinities between them. Rather than using individual client scores or pairwise similarities directly, FedCRF treats them as evidence for jointly inferring these latent reliabilities over this graph, so that evidence about one client can affect the assessment of related clients. The resulting posterior reliabilities are then used to correct the standard FedAvg aggregation weights.

We instantiate this formulation using two complementary sources of evidence. \emph{Geometric potentials} require only the returned client models and updates: individual update statistics provide unary evidence and update similarities provide pairwise affinities. \emph{Functional potentials} instead use a small server-side proxy dataset: changes in proxy loss provide unary evidence, while agreement between client predictions defines pairwise affinity. The two sources can also be combined.

Crucially, the inferred client reliabilities and affinities provide a common representation that can be reused across different FL aggregation settings. For global aggregation and Byzantine robustness, inferred reliabilities adjust the influence of client contributions. For clustered FL, the affinity structure is used to identify groups of related clients, within which reliability-aware aggregation is performed. For personalized FL, affinities and reliabilities jointly determine client-specific model mixtures. 
FedCRF therefore provides a shared inference layer for interpreting client contributions, rather than requiring a separate client-assessment mechanism for each aggregation setting.

Our contributions are threefold. 
\textbf{First}, we formulate server-side aggregation as structured probabilistic inference over a client graph, 
jointly inferring latent client reliabilities from individual evidence and pairwise relationships. 
\textbf{Second}, we introduce geometric and functional constructions of the CRF potentials and show that the same inferred reliability and affinity representation supports \emph{global aggregation, Byzantine robustness, clustered FL, and personalized FL} with a single set of hyperparameters. To the best of our knowledge, FedCRF is the first method to address all four settings through one shared client-assessment mechanism, whereas prior work targets one or two of them with dedicated designs.
\textbf{Third}, we evaluate FedCRF on \emph{CIFAR-10, FEMNIST, and Fed-ISIC2019} under distribution shifts, heterogeneous client quality, partial participation, malicious updates, and group- and client-level heterogeneity, comparing against specialized methods for the corresponding settings. Our implementation, including the configuration of every experiment, is available at \url{https://anonymous.4open.science/r/fedcrf-code-0044}.

\begin{figure}[t]
    \centering
    \vspace{-5pt}
    \includegraphics[width=\textwidth]{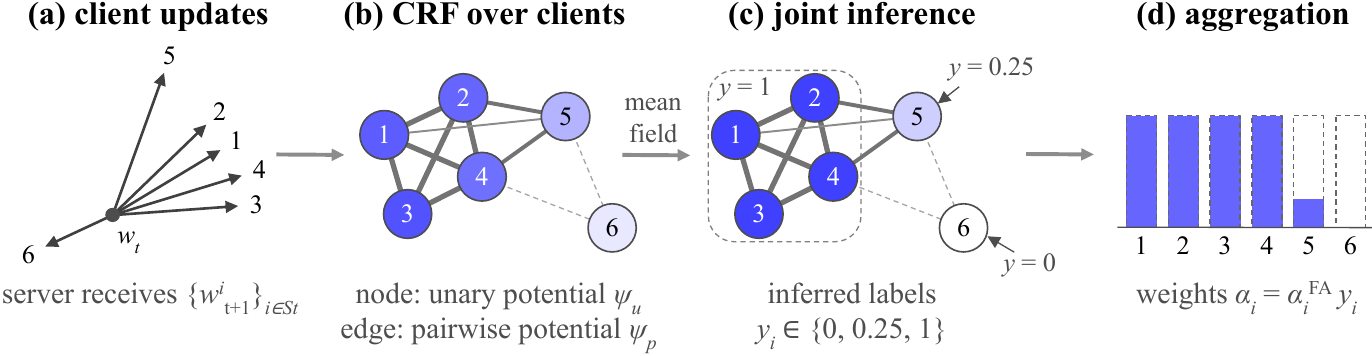}
    \captionsetup{skip=7pt}
    \caption{Illustration of the proposed method. (a)~Client updates provide evidence for (b)~a CRF with unary reliability potentials and pairwise client affinities. (c)~Joint inference assigns labels that retain, down-weight, or exclude contributions. (d)~These labels rescale the FedAvg weights.}\vspace{-8pt}
    \label{fig:overview}
\end{figure}

\vspace{-2pt}\section{Background}
\label{sec:background}

\subsection{Conditional Random Fields}
\label{sec:bg-crf}

Conditional random fields (CRFs) are probabilistic graphical models that define a conditional distribution over structured data given observations \citep{lafferty2001conditional}. Let $G=(V,E)$ be a graph with $|V|=N$ nodes and edges $E$. For each node $i\in V$, we observe an input $x_i\in\mathcal{X}$ and associate a random variable $y_i\in\mathcal{Y}$. A CRF defines the conditional distribution
\begin{equation}
p(\boldsymbol{y}\mid\boldsymbol{x})
=
\frac{1}{Z(\boldsymbol{x})}
\exp\big(-E(\boldsymbol{y}\mid\boldsymbol{x})\big),
\quad 
E(\boldsymbol{y}\mid\boldsymbol{x})
=
\sum_i \psi_u(y_i,x_i)
+
\sum_{i<j}\psi_p(y_i,y_j,x_i,x_j),
\end{equation}
where $E(\boldsymbol{y}\mid\boldsymbol{x})$ is the energy of a configuration, and $Z(\boldsymbol{x})$ is the normalizing partition function. 

The \emph{unary potential} $\psi_u(y_i,x_i)$ measures the cost of assigning label $y_i$ to node $i$ given its observation $x_i$. It is often constructed as a negative log-probability, so labels better supported by individual evidence receive lower energy. The \emph{pairwise potential} $\psi_p(y_i,y_j,x_i,x_j)$ encodes interactions between nodes. A common parameterization is
\begin{equation}
\psi_p(y_i,y_j,x_i,x_j)
=
\mu(y_i,y_j)\sum_{m=1}^{M}a_m k_m(x_i,x_j),
\end{equation}
where $k_m(x_i,x_j)\geq 0$ are observation-dependent affinities, $a_m\geq 0$ are their weights, and $\mu(y_i,y_j)$ specifies the compatibility of a label pair.

The \emph{maximum a posteriori} (MAP) configuration minimizes the energy $\boldsymbol{y}^{*}
=
\arg\max_{\boldsymbol{y}}p(\boldsymbol{y}\mid\boldsymbol{x})
=
\arg\min_{\boldsymbol{y}}E(\boldsymbol{y}\mid\boldsymbol{x})$. For general potentials on dense graphs, exact MAP inference and exact marginal computation are computationally intractable. A common approach to approximate marginal inference is the \emph{mean-field approximation} \citep{wainwright2008graphical,krahenbuhl2011efficient}, which introduces a node-factorized distribution $q(\boldsymbol{y}\mid\boldsymbol{x})
=
\prod_i q_i(y_i\mid\boldsymbol{x})
$
and minimizes $\mathrm{KL}(q|p)$.

\subsection{Federated Learning}
\label{sec:background-fl}

Consider $K$ clients, where client $k$ owns a local dataset $\mathcal{D}_k$ containing $n_k$ examples, with $n=\sum_{k=1}^{K} n_k$ examples in total. In the standard empirical FL formulation, the goal is to collaboratively learn a shared model parameter vector $w\in\mathbb{R}^{d}$ across clients by minimizing
\begin{equation}
\min_w F(w)
=
\sum_{k=1}^{K} p_k F_k(w),
\qquad
p_k=\frac{n_k}{n},
\qquad
F_k(w)
=
\frac{1}{n_k}
\sum_{(x,y)\in\mathcal{D}_k}
\ell(w;x,y).
\end{equation}
Other client-weighted objectives are possible, but weighting clients by their number of local examples is the standard choice underlying FedAvg \citep{mcmahan2017communication}.

FedAvg optimizes this objective through repeated rounds of local training and server aggregation. At communication round $t$, the server broadcasts the current model $w_t$ to a subset of participating clients $S_t$. Each participating client initializes from $w_t$, performs local optimization on its dataset, and returns a locally trained model $w_{t+1}^{k}$. The server then aggregates the returned models as
\begin{equation}
w_{t+1}
=
\sum_{k\in S_t}
\alpha_k^{\mathrm{FA}} w_{t+1}^{k},
\qquad
\alpha_k^{\mathrm{FA}}
=
\frac{n_k}{\sum_{j\in S_t}n_j}.
\label{eq:fedavg}
\end{equation}

\paragraph{Federated aggregation settings.}
FL methods differ in how client contributions are combined. \emph{Global aggregation} learns a single model shared across clients, while \emph{clustered FL} addresses structured heterogeneity by learning shared models for groups of related clients \citep{sattler2019cfl,ghosh2020ifca,ruan2022fedsoft,guo2024fedrc}. \emph{Personalized FL} operates at a finer granularity, allowing client-specific models while retaining collaboration across clients \citep{dinh2020pfedme,li2021ditto,zhang2021fedfomo}. These settings mainly differ in how they accommodate client heterogeneity and the desired degree of model sharing. Orthogonally, \emph{Byzantine-robust aggregation} considers that some clients may submit arbitrary or adversarial updates and seeks to limit their influence \citep{blanchard2017machine,yin2018byzantine,pillutla2022robust, fenoglioFBP2024h}. Thus, disagreement among client contributions may motivate different strategies, such as weighting, grouping, personalization, or robust aggregation, which are typically addressed by specialized methods. We discuss these approaches in detail in \cref{sec:related}.

\section{Method}
\label{sec:method}

We introduce \textbf{FedCRF}, a method that formulates FL server-side aggregation as structured probabilistic inference over client contributions. Starting from FedAvg, we retain local training and use a CRF to infer correction coefficients for the aggregation weights.

\subsection{CRF for Federated Learning}
\label{sec:method-overview}

At each round $t$, we represent the participating clients as a fully connected graph $G=(V,E)$, with one node per client $i\in S_t$. Each node has a latent label $y_i\in\mathcal{Y}$ that specifies how to adjust its contribution: retain its unnormalized FedAvg weight, reduce it, or exclude it, respectively. In aggregation, we use the posterior expectation of this label as the correction coefficient (details in \cref{sec:method-inference}). The observation $x_i$ contains evidence from the client's update $u_i=w_{t+1}^{i}-w_t$ and, when available, its model's predictions on a server-side proxy dataset. We omit the round index below for simplicity.

To instantiate the CRF, we define the unary potential, pairwise potential, and label compatibility function as
\begin{equation}
    \psi_u(y_i,x_i)=-\log\pi_i(y_i),
    \quad
    \psi_p(y_i,y_j,x_i,x_j)=\lambda A_{ij}\mu(y_i,y_j),
    \quad
    \mu(y_i,y_j)=(y_i-y_j)^2.
    \label{eq:pairwise}
\end{equation}
Here, $\pi_i$ is a prior distribution over the client's labels, $A_{ij}\geq0$ is a symmetric affinity between clients, and $\lambda\geq0$ controls the strength of the pairwise term. The squared label difference makes disagreement more costly for strongly connected clients and distinguishes partial down-weighting from complete exclusion.

We construct the unary prior from an individual reliability score $\rho_i\in(0,1)$:
\begin{equation}
    \pi_i(y)=
    \frac{b_i(y)}{\sum_{\ell\in\mathcal{Y}}b_i(\ell)},
    \qquad
    b_i(y)=\rho_i^{\,y}\,(1-\rho_i)^{\,1-y}.
    \label{eq:label-init}
\end{equation}
The endpoint scores are $b_i(1)=\rho_i$ and $b_i(0)=1-\rho_i$, while $b_i(0.25)=\rho_i^{1/4}(1-\rho_i)^{3/4}$ is between them. Thus, larger reliability scores favor retaining a contribution, and smaller scores favor excluding it, while at $\rho_i=\tfrac12$ the prior is uniform over the three labels. We next specify how client evidence determines $\rho_i$ and $A_{ij}$.

\subsection{Geometric and functional potentials}
\label{sec:method-evidence}

\paragraph{Geometric evidence: $\mathrm{FedCRF}_{G}$.}
We use client updates $u_i=w_{t+1}^{i}-w_t$ and their coordinate-wise median $\tilde{u}=\mathrm{coordmed}_{j\in S_t}u_j$, requiring no server-side data. For each client, we define cosine alignment, normalized distance, and norm anomaly as
\begin{equation}
    c_i=\cos(u_i,\tilde{u}),
    \qquad
    d_i=\frac{\|u_i-\tilde{u}\|_2}{\operatorname{med}_j\|u_j-\tilde{u}\|_2},
    \qquad
    h_i=\frac{\big|\|u_i\|_2-\operatorname{med}_j\|u_j\|_2\big|}
    {\operatorname{MAD}_j(\|u_j\|_2)}.
\end{equation}
Here, $\operatorname{MAD}_j(z_j)=\operatorname{med}_j|z_j-\operatorname{med}_k z_k|$, all medians are over participating clients. The reliability and affinities are
\begin{equation}
    \rho_i^{G}=\sigma(2c_i-d_i-0.5h_i),
    \qquad
    A_{ij}^{G}=\exp\left(-\frac{1-\cos(u_i,u_j)}{\theta_G}\right)
    \min(\rho_i^{G},\rho_j^{G}),
    \label{eq:rho-geo}
\end{equation}
where $\sigma$ is the logistic sigmoid, $\theta_G>0$, and $A_{ii}^{G}=0$. The affinity combines directional agreement with a reliability gate that limits strong connections within unreliable clients and between reliable and unreliable ones.

\paragraph{Functional evidence: $\mathrm{FedCRF}_{F}$.}
This variant assumes a small proxy dataset $\mathcal{D}_{\mathrm{prox}}$ at the server and constructs evidence from model performance and predictive agreement. For each client, we measure the improvement in proxy loss,
\begin{equation}
    \delta_i
    =
    \mathcal{L}_{\mathrm{prox}}(w_t)
    -
    \mathcal{L}_{\mathrm{prox}}(w_{t+1}^{i}),
    \qquad
    \rho_i^{F}=\sigma(\hat{\delta}_i/\tau),
    \label{eq:rho-functional}
\end{equation}
where $\hat{\delta}_i$ is standardized using the median and median absolute deviation across participating clients, and $\tau>0$ is a temperature.

Let $p_i(\cdot\mid x)$ denote the predictive distribution of $w_{t+1}^{i}$. We compute the empirical mean symmetric divergence on proxy inputs and the corresponding affinity:
\begin{equation}
    D_{ij}
    =
    \frac{1}{2}
    \mathbb{E}_{x\sim\mathcal{D}_{\mathrm{prox}}}
    \left[
        \mathrm{KL}(p_i\|p_j)
        +
        \mathrm{KL}(p_j\|p_i)
    \right],
    \qquad
    A_{ij}^{F}
    =
    \exp(-D_{ij}/\theta_F)
    \min(\rho_i^{F},\rho_j^{F}),
    \label{eq:affinity-val}
\end{equation}
with zero diagonal. Thus, unary evidence measures relative improvement in proxy performance, while pairwise evidence measures agreement between predictions.

\paragraph{Mixed evidence: $\mathrm{FedCRF}_{G+F}$.}
We also combine the two constructions:
\begin{equation}
    \rho_i^{G+F}=\frac{\rho_i^{G} + \rho_i^{F}}{2},
    \qquad
    A_{ij}^{G+F}=\sqrt{A_{ij}^{G}A_{ij}^{F}},
    \quad i\neq j.
    \label{eq:hybrid}
\end{equation}

\subsection{Inference to aggregation}
\label{sec:method-inference}

We approximate the conditional distribution using the mean-field updates from \cref{sec:bg-crf}. Starting from $q_i^{(0)}(y)=\pi_i(y)$, we perform $R$ mean-field updates:
\begin{equation}
    q_i^{(r+1)}(y)=
    \frac{
        \exp\!\left[-\psi_u(y,x_i)-\lambda\sum_{j\neq i}A_{ij}
        \sum_{\ell\in\mathcal{Y}}q_j^{(r)}(\ell)(y-\ell)^2\right]
    }{
        \sum_{z\in\mathcal{Y}}\exp\!\left[-\psi_u(z,x_i)-\lambda\sum_{j\neq i}A_{ij}
        \sum_{\ell\in\mathcal{Y}}q_j^{(r)}(\ell)(z-\ell)^2\right]
    }.
    \label{eq:fedcrf-inference}
\end{equation}
The posterior expectation $\bar{y}_i$ provides the correction coefficient for client $i$. Using the FedAvg weights from \cref{eq:fedavg}, the server computes
\begin{equation}
    \bar{y}_i=\sum_{y\in\mathcal{Y}}y\,q_i^{(R)}(y),
    \qquad
    \alpha_i=\frac{\alpha_i^{\mathrm{FA}}\bar{y}_i}
    {\sum_{j\in S_t}\alpha_j^{\mathrm{FA}}\bar{y}_j},
    \qquad
    w_{t+1}=\sum_{i\in S_t}\alpha_i w_{t+1}^{i}.
    \label{eq:crf-weight}
\end{equation}
FedCRF therefore adjusts sample-count weights using jointly inferred reliabilities. Equal positive coefficients recover FedAvg.

\subsection{Extensions to Group-Wise and Client-Specific Aggregation}
\label{sec:beyond}

\paragraph{Clustered FL.}
To adapt for Clustered FL setting, we use client similarities before reliability gating to get locally scaled affinities $\tilde{A}_{ij}$ \citep{zelnikmanor2004selftuning}. Then we use spectral clustering to identify client groups, with their number specified or selected by the eigengap. Each group then trains a shared model using FedCRF aggregation among its members.

\paragraph{Personalized FL.}
When clients do not form clear groups, the goal can instead be to learn a separate model for each client. The server still uses pairwise affinities to share model weights between similar clients, but each client receives its own combination of the returned models. After local training from personalized models $w_i^{\mathrm{pers}}$, the server updates the distance graph, recomputes the locally scaled affinities, and mixes the returned models as
\begin{equation}
    w_i^{\mathrm{pers}}\leftarrow\sum_j W_{ij}w_{t+1}^{j},
    \qquad
    W_{ij}
    =
    \frac{B_{ij}\bar{y}_j}
    {\sum_{\ell}B_{i\ell}\bar{y}_{\ell}},
    \label{eq:crfp}
\end{equation}
where $B_{ij}=\tilde{A}_{ij}^{1/T}$ for $j\neq i$ and $B_{ii}=s$. Temperature $T>0$ controls neighbor selectivity, and $s>0$ controls self-weight. Each client's mixing weights therefore reflect both its affinities to other clients and their inferred reliability.

\section{Experiments}
\label{sec:experiments}

\subsection{Setup}
\label{sec:setup}
\paragraph{Datasets and models.}
We evaluate controlled heterogeneity on CIFAR-10 (ResNet-9) and natural federation on FEMNIST and Fed-ISIC2019. ANDA \citep{liFederatedLearningProfile2025b} partitions CIFAR-10 by label skew $P(y)$, feature skew $P(x)$, concept shift $P(y\mid x)$, or class-conditional feature shift $P(x\mid y)$, at medium and high severity, alongside independent and identically distributed (IID) data. Its \emph{strict} variants provide known client groups for evaluating clustering. FEMNIST \citep{caldas2018leaf} uses writers as clients (62 classes, LEAF CNN). Fed-ISIC2019 from FLamby \citep{flamby2022} uses an ImageNet-pretrained EfficientNet-B0 for 8 classes; its 6 acquisition centres are split into sub-clients, with centre identity as the reference grouping.

\paragraph{Training setup.}
We implement FedCRF and all baselines in PyTorch in a single code base, so that all methods share the data partitions, local training, attacks, and evaluation. Unless stated otherwise, we use 20 clients with full participation, 2 local epochs, batch size 64, and SGD with learning rate $0.01$ and momentum $0.9$. Training runs for at most 300 rounds, with early stopping after 5 rounds without improvement in sample-weighted client validation loss. Each client's data is split 80/10/10 for training, validation, and testing; results use 3 seeds. The proxy set contains 10 examples per class from the official test split, excluded from evaluation. Our main metric is sample-weighted client test accuracy; we additionally report official test accuracy on the remaining benchmark test set. Default hyperparameter values are $(\lambda,\theta_G,\theta_F,\tau)=(0.5,0.5,1,1)$, $\mathcal{Y}=\{0,0.25,1\}$, $R=5$, and personalization self-weight $s=10$. We use the same set of hyperparameters for all the datasets and settings.  FedCRF is not sensitive to hyperparameter choices and we did not tune them for any of the tables: across the 21 settings of the sensitivity grid, validation accuracy varies by at most 1.6 points (see Appendix \cref{app:ablations}). Tables mark the best mean performance in bold and the second best underlined. Extra details are available in Appendix \cref{app:setup}, we also include everything required to reproduce the results in our anonymized repository.

\paragraph{Baselines.}
We compare global aggregation with FedAvg, FedProx \citep{li2020federated}, FedNova \citep{wang2020fednova}, FedLAW \citep{li2023fedlaw}, and FedLWS \citep{shi2025fedlws}. 
Robustness baselines include coordinate-wise median, trimmed mean \citep{yin2018byzantine}, Krum and Multi-Krum \citep{blanchard2017machine}, RFA \citep{pillutla2022robust}, geometric median, Fed-NGA \citep{zuoEfficientFederatedLearning2025}, FLTrust \citep{cao2021fltrust}, and Zeno \citep{xie2019zeno}; the latter two use the same proxy set as FedCRF. For clustering and personalization, we include local training, FedAvg with local fine-tuning, Ditto \citep{li2021ditto}, pFedGraph \citep{ye2023pfedgraph}, IFCA \citep{ghosh2020ifca}, CFL \citep{sattler2019cfl}, FedSoft \citep{ruan2022fedsoft}, FedRC \citep{guo2024fedrc}, FedFomo \citep{zhang2021fedfomo}, and FLUX \citep{fenoglioFLUX2025c}. A 
detailed description of the baselines is given in \cref{app:baselines}.

\subsection{Global aggregation}
\label{sec:exp-global}


\paragraph{Heterogeneous clients.}
\cref{tab:quality} reports results for federated training with heterogeneous clients. For CIFAR-10, we use ANDA \citep{liFederatedLearningProfile2025b} to generate non-IID client distributions, while FEMNIST and ISIC2019 provide naturally heterogeneous data. In case of no data corruption (see \emph{clean} columns in \cref{tab:quality}), all three versions of the proposed approach generally perform on par with the best baselines, and even outperform them slightly in the most complex case of ISIC2019. 

To test the methods in a more difficult setup, we corrupt the training data of 30\% of clients by injecting standard Gaussian noise into the input images, leaving validation and test data clean (see \emph{corrupted} in \cref{tab:quality}). Across the six clean and corrupted settings, \crfg{} and \crfh{} have the highest mean accuracy (73.6\% and 73.0\%, versus 71.9\% for the best baseline). The clearest gains occur on corrupted CIFAR-10 and Fed-ISIC2019, where all three variants outperform all the baselines, led by \crfv{} on CIFAR-10 at 82.7\% versus 78.1\% for FedLWS. On Fed-ISIC2019, \crfg{} gets the best result and reaches 62.0\% versus 53.5\% for FedLWS, while functional and mixed evidence offer smaller gains but still outperform the best baseline. We further show in Appendix \cref{app:diagnostics} how each evidence mode suppresses corrupted clients.

\begin{table}[t]
\centering
\caption{Global aggregation under heterogeneous clients. In \emph{corrupted} case, 30\% of the clients train on data corrupted with Gaussian noise. Client test accuracy (\%, $\uparrow$), 3 seeds.}
\label{tab:quality}
\small
\setlength{\tabcolsep}{4pt}
\resizebox{\ifdim\width>\textwidth\textwidth\else\width\fi}{!}{%
\begin{tabular}{lccccccc}
\toprule
Method & \multicolumn{2}{c}{CIFAR-10, $P(y)$ medium} & \multicolumn{2}{c}{FEMNIST (20 writers)} & \multicolumn{2}{c}{Fed-ISIC2019} &   \\
\cmidrule(lr){2-3} \cmidrule(lr){4-5} \cmidrule(lr){6-7}
 & clean & corrupted & clean & corrupted & clean & corrupted & avg \\
\midrule
\multicolumn{8}{l}{\emph{Baselines}} \\
FedAvg & 84.5{\scriptsize$\pm$0.3} & 77.5{\scriptsize$\pm$0.8} & \textbf{74.0{\scriptsize$\pm$2.9}} & 72.4{\scriptsize$\pm$4.6} & 68.3{\scriptsize$\pm$0.7} & 53.3{\scriptsize$\pm$0.9} & 71.7 \\
FedProx & 84.7{\scriptsize$\pm$0.7} & 77.3{\scriptsize$\pm$1.1} & 74.0{\scriptsize$\pm$2.7} & \underline{72.7{\scriptsize$\pm$5.1}} & 68.3{\scriptsize$\pm$0.4} & 53.3{\scriptsize$\pm$1.0} & 71.7 \\
FedNova & \underline{84.8{\scriptsize$\pm$0.4}} & 77.4{\scriptsize$\pm$1.1} & 73.6{\scriptsize$\pm$1.6} & 72.1{\scriptsize$\pm$5.3} & 67.9{\scriptsize$\pm$1.0} & 53.2{\scriptsize$\pm$0.9} & 71.5 \\
FedLWS & \textbf{85.3{\scriptsize$\pm$0.9}} & 78.1{\scriptsize$\pm$0.9} & \underline{74.0{\scriptsize$\pm$2.8}} & 72.5{\scriptsize$\pm$5.1} & 68.2{\scriptsize$\pm$1.1} & 53.5{\scriptsize$\pm$0.9} & 71.9 \\
FedLAW & 77.4{\scriptsize$\pm$6.8} & 72.7{\scriptsize$\pm$0.6} & 5.0{\scriptsize$\pm$1.1} & 5.6{\scriptsize$\pm$0.8} & 49.1{\scriptsize$\pm$0.8} & 49.1{\scriptsize$\pm$0.8} & 43.1 \\
\midrule
\multicolumn{8}{l}{\emph{Ours}} \\
\crfg{} & 84.6{\scriptsize$\pm$0.9} & 80.8{\scriptsize$\pm$0.7} & 73.3{\scriptsize$\pm$4.5} & 71.4{\scriptsize$\pm$5.1} & \textbf{69.7{\scriptsize$\pm$0.9}} & \textbf{62.0{\scriptsize$\pm$3.4}} & \textbf{73.6} \\
\crfv{} & 84.3{\scriptsize$\pm$1.2} & \textbf{82.7{\scriptsize$\pm$1.0}} & 71.7{\scriptsize$\pm$5.3} & 71.1{\scriptsize$\pm$3.4} & \underline{69.2{\scriptsize$\pm$0.5}} & 54.4{\scriptsize$\pm$3.4} & 72.2 \\
\crfh{} & 84.7{\scriptsize$\pm$0.5} & \underline{82.0{\scriptsize$\pm$1.8}} & 73.7{\scriptsize$\pm$4.3} & \textbf{72.9{\scriptsize$\pm$3.3}} & 69.1{\scriptsize$\pm$0.5} & \underline{55.5{\scriptsize$\pm$2.4}} & \underline{73.0} \\
\bottomrule
\end{tabular}}
\end{table}

\paragraph{Partial participation.}
\cref{tab:participation} shows results under partial client participation, where at each round only a small random cohort participates. Our method with functional and mixed evidence achieves the highest averages across the six settings in \cref{tab:participation}: 65.5\% and 65.2\%, versus 63.5\% for FedProx. On CIFAR-10 $P(y)$ medium with 50 clients and 20\% participation, \crfv{} reaches 71.6\%, compared with 54.1\% for FedAvg and 70.2\% for FedLAW, the strongest baseline in this case. This advantage is not uniform: with 100 clients and 10\% participation, FedLAW leads at 61.2\%, while all other methods remain below 33\%. However, this strong performance does not transfer consistently across datasets, as FedLAW drops to 5.2\% and 6.2\% on FEMNIST and to 57.6\% and 54.2\% on Fed-ISIC2019 under 20\% and 10\% participation, respectively, while the other methods remain around 75--78\% and 67--68.5\%. On Fed-ISIC2019, whose six centres we split into 50 or 100 sub-clients, \crfv{} and \crfh{} are the most accurate 
with 100 clients (67.9\%), and \crfh{} is within 0.2 points of the best method with 50 clients (68.3\% versus 68.5\% for FedProx). 

\begin{table}[t]
\centering
\caption{Partial participation: client test accuracy (\%, $\uparrow$) with $K$ clients and the given participation rate per round (10-client cohorts), 3 seeds. Fed-ISIC2019 splits its six centres into $K$ sub-clients in proportion to their size.}
\label{tab:participation}
\small
\setlength{\tabcolsep}{4pt}
\resizebox{\ifdim\width>\textwidth\textwidth\else\width\fi}{!}{%
\begin{tabular}{lccccccc}
\toprule
Method & \multicolumn{2}{c}{CIFAR-10, $P(y)$ medium} & \multicolumn{2}{c}{FEMNIST} & \multicolumn{2}{c}{Fed-ISIC2019} &   \\
\cmidrule(lr){2-3} \cmidrule(lr){4-5} \cmidrule(lr){6-7}
 & $K{=}50$, 20\% & $K{=}100$, 10\% & $K{=}50$, 20\% & $K{=}100$, 10\% & $K{=}50$, 20\% & $K{=}100$, 10\% & avg \\
\midrule
\multicolumn{8}{l}{\emph{Baselines}} \\
FedAvg & 54.1{\scriptsize$\pm$10.8} & 31.8{\scriptsize$\pm$4.7} & 77.1{\scriptsize$\pm$3.1} & \textbf{78.2{\scriptsize$\pm$1.6}} & 68.1{\scriptsize$\pm$2.2} & 67.4{\scriptsize$\pm$1.1} & 62.8 \\
FedProx & 64.1{\scriptsize$\pm$18.6} & 25.6{\scriptsize$\pm$11.2} & \underline{77.8{\scriptsize$\pm$2.9}} & 78.0{\scriptsize$\pm$1.4} & \textbf{68.5{\scriptsize$\pm$1.8}} & 66.8{\scriptsize$\pm$1.7} & 63.5 \\
FedNova & 63.4{\scriptsize$\pm$16.7} & 18.7{\scriptsize$\pm$2.5} & 75.5{\scriptsize$\pm$1.5} & 76.7{\scriptsize$\pm$1.1} & 67.7{\scriptsize$\pm$1.8} & 67.3{\scriptsize$\pm$1.4} & 61.6 \\
FedLWS & 68.1{\scriptsize$\pm$4.4} & 17.0{\scriptsize$\pm$5.2} & \textbf{77.8{\scriptsize$\pm$2.7}} & \underline{78.0{\scriptsize$\pm$1.5}} & 67.6{\scriptsize$\pm$1.7} & 66.9{\scriptsize$\pm$1.0} & 62.6 \\
FedLAW & 70.2{\scriptsize$\pm$9.4} & \textbf{61.2{\scriptsize$\pm$3.0}} & 5.2{\scriptsize$\pm$0.7} & 6.2{\scriptsize$\pm$1.0} & 57.6{\scriptsize$\pm$1.1} & 54.2{\scriptsize$\pm$5.9} & 42.4 \\
\midrule
\multicolumn{8}{l}{\emph{Ours}} \\
\crfg{} & 57.4{\scriptsize$\pm$14.1} & 24.7{\scriptsize$\pm$7.9} & 75.8{\scriptsize$\pm$1.6} & 77.9{\scriptsize$\pm$1.1} & 68.1{\scriptsize$\pm$1.3} & 67.6{\scriptsize$\pm$0.6} & 61.9 \\
\crfv{} & \textbf{71.6{\scriptsize$\pm$3.5}} & \underline{32.9{\scriptsize$\pm$2.2}} & 75.3{\scriptsize$\pm$3.2} & 77.5{\scriptsize$\pm$1.4} & 67.9{\scriptsize$\pm$1.2} & \textbf{67.9{\scriptsize$\pm$0.4}} & \textbf{65.5} \\
\crfh{} & \underline{71.0{\scriptsize$\pm$14.2}} & 30.9{\scriptsize$\pm$3.2} & 75.7{\scriptsize$\pm$2.3} & 77.2{\scriptsize$\pm$0.8} & \underline{68.3{\scriptsize$\pm$1.3}} & \underline{67.9{\scriptsize$\pm$0.7}} & \underline{65.2} \\
\bottomrule
\end{tabular}}
\end{table}

\subsection{Byzantine robustness}
\label{sec:exp-robust}
In \cref{tab:robust} we evaluate our method under five attacks with 20\% fixed malicious clients: sign flipping, ALIE \citep{baruch2019alie}, targeted label flipping, model replacement \citep{bagdasaryan2020backdoor}, and Neurotoxin \citep{zhang2022neurotoxin}. We report clean accuracy and, for the targeted attacks, attack success rate (ASR). 

\begin{table}[t]
\centering
\caption{Byzantine robustness on CIFAR-10 ($P(y)$ medium, 20\% fixed malicious clients, 3 seeds): clean accuracy (\%) under five attacks and attack success rate (ASR, \%) for the targeted attacks.}
\label{tab:robust}
\small
\setlength{\tabcolsep}{4pt}
\resizebox{\ifdim\width>\textwidth\textwidth\else\width\fi}{!}{%
\begin{tabular}{lcccccccccc}
\toprule
Method & \multicolumn{6}{c}{Clean accuracy under attack $\uparrow$} & \multicolumn{4}{c}{Attack success rate $\downarrow$} \\
\cmidrule(lr){2-7} \cmidrule(lr){8-11}
 & sign flip & ALIE & targeted LF & model repl. & Neurotoxin & avg & targeted LF & model repl. & Neurotoxin & avg \\
\midrule
\multicolumn{11}{l}{\emph{Robust aggregation}} \\
Coord.\ median & 78.4{\scriptsize$\pm$1.8} & 54.0{\scriptsize$\pm$13.5} & 83.2{\scriptsize$\pm$1.3} & \underline{84.8{\scriptsize$\pm$0.4}} & \textbf{85.0{\scriptsize$\pm$1.0}} & 77.1 & 3.8{\scriptsize$\pm$1.3} & 94.9{\scriptsize$\pm$0.2} & 93.0{\scriptsize$\pm$1.9} & 63.9 \\
Trimmed mean & 73.9{\scriptsize$\pm$3.5} & 70.6{\scriptsize$\pm$1.3} & 82.8{\scriptsize$\pm$1.3} & \textbf{85.1{\scriptsize$\pm$0.9}} & \underline{84.9{\scriptsize$\pm$0.3}} & 79.5 & 9.9{\scriptsize$\pm$4.1} & 95.7{\scriptsize$\pm$0.7} & 94.7{\scriptsize$\pm$0.9} & 66.8 \\
Krum & 46.6{\scriptsize$\pm$24.1} & 27.6{\scriptsize$\pm$9.3} & 70.2{\scriptsize$\pm$4.5} & 51.6{\scriptsize$\pm$1.8} & 51.6{\scriptsize$\pm$1.8} & 49.5 & 1.9{\scriptsize$\pm$0.7} & 6.5{\scriptsize$\pm$4.2} & 6.5{\scriptsize$\pm$4.2} & 5.0 \\
Multi-Krum & \textbf{83.9{\scriptsize$\pm$0.7}} & 65.5{\scriptsize$\pm$4.9} & 82.5{\scriptsize$\pm$1.5} & 83.4{\scriptsize$\pm$1.0} & 83.4{\scriptsize$\pm$1.0} & 79.8 & \underline{0.7{\scriptsize$\pm$0.4}} & \textbf{1.5{\scriptsize$\pm$0.4}} & \underline{1.5{\scriptsize$\pm$0.4}} & \textbf{1.3} \\
RFA & 81.8{\scriptsize$\pm$0.4} & 68.5{\scriptsize$\pm$2.6} & 80.9{\scriptsize$\pm$4.0} & 83.8{\scriptsize$\pm$0.8} & 84.6{\scriptsize$\pm$0.9} & 79.9 & 6.5{\scriptsize$\pm$2.4} & 92.4{\scriptsize$\pm$1.7} & 89.4{\scriptsize$\pm$1.4} & 62.8 \\
Geom.\ median & 81.8{\scriptsize$\pm$0.4} & 68.5{\scriptsize$\pm$2.6} & 80.9{\scriptsize$\pm$4.0} & 83.8{\scriptsize$\pm$0.8} & 84.6{\scriptsize$\pm$0.9} & 79.9 & 6.5{\scriptsize$\pm$2.4} & 92.4{\scriptsize$\pm$1.7} & 89.4{\scriptsize$\pm$1.4} & 62.8 \\
Fed-NGA & 9.9{\scriptsize$\pm$0.1} & 65.2{\scriptsize$\pm$3.3} & 77.0{\scriptsize$\pm$0.9} & 57.4{\scriptsize$\pm$39.9} & 10.4{\scriptsize$\pm$0.9} & 44.0 & 13.6{\scriptsize$\pm$4.5} & 60.1{\scriptsize$\pm$52.1} & \textbf{0.0{\scriptsize$\pm$0.0}} & 24.6 \\
\midrule
\multicolumn{11}{l}{\emph{With server data}} \\
FLTrust & 73.1{\scriptsize$\pm$0.1} & 76.4{\scriptsize$\pm$1.4} & 77.4{\scriptsize$\pm$0.8} & 77.7{\scriptsize$\pm$1.4} & 77.6{\scriptsize$\pm$1.4} & 76.5 & 3.4{\scriptsize$\pm$1.0} & 73.4{\scriptsize$\pm$3.0} & 2.3{\scriptsize$\pm$0.6} & 26.4 \\
Zeno & \underline{83.4{\scriptsize$\pm$0.5}} & \textbf{83.7{\scriptsize$\pm$0.9}} & 82.8{\scriptsize$\pm$0.5} & 82.8{\scriptsize$\pm$0.3} & 83.0{\scriptsize$\pm$0.2} & \underline{83.1} & 0.8{\scriptsize$\pm$0.3} & \underline{1.9{\scriptsize$\pm$0.3}} & 1.6{\scriptsize$\pm$0.6} & \underline{1.4} \\
\midrule
\multicolumn{11}{l}{\emph{Ours}} \\
\crfg{} & 81.5{\scriptsize$\pm$0.7} & 76.7{\scriptsize$\pm$1.6} & 82.8{\scriptsize$\pm$1.1} & 82.7{\scriptsize$\pm$0.7} & 83.3{\scriptsize$\pm$0.6} & 81.4 & 4.0{\scriptsize$\pm$2.9} & 32.6{\scriptsize$\pm$27.2} & 8.0{\scriptsize$\pm$3.4} & 14.9 \\
\crfv{} & 83.0{\scriptsize$\pm$0.6} & \underline{83.6{\scriptsize$\pm$0.7}} & \textbf{83.7{\scriptsize$\pm$1.0}} & 83.4{\scriptsize$\pm$0.2} & 83.3{\scriptsize$\pm$1.9} & \textbf{83.4} & \textbf{0.6{\scriptsize$\pm$0.1}} & 2.5{\scriptsize$\pm$0.4} & 1.9{\scriptsize$\pm$0.5} & 1.7 \\
\crfh{} & 82.0{\scriptsize$\pm$0.3} & 77.9{\scriptsize$\pm$1.3} & \underline{83.5{\scriptsize$\pm$0.4}} & 82.9{\scriptsize$\pm$1.9} & 83.4{\scriptsize$\pm$1.0} & 81.9 & 1.0{\scriptsize$\pm$0.2} & 12.5{\scriptsize$\pm$16.1} & 6.2{\scriptsize$\pm$6.1} & 6.6 \\
\bottomrule
\end{tabular}}
\end{table}

Functional evidence provides the strongest overall results, while other variants still perform better than most of the baselines on average (except for Zeno which gets second best score). \crfv{} averages 83.4\% clean accuracy across five attacks, on par with Zeno's 83.1\% and above the best baseline without server data (79.9\%; \cref{tab:robust}). 
Under ALIE, \crfv{} reaches 83.6\%, essentially matching Zeno at 83.7\% with the same proxy set, compared with 76.7\% for geometric evidence and 70.6\% for trimmed mean.

Under model replacement and Neurotoxin, \crfv{} combines 83.3--83.4\% clean accuracy with ASR of 2.5\% and 1.9\%. Coordinate-wise median, trimmed mean, RFA, and geometric median retain high clean accuracy but exceed 89\% ASR. Overall, FedCRF achieves a strong balance between high accuracy and low ASR.
Reliability diagnostics further confirm that malicious clients are assigned substantially lower inferred reliability over training rounds (Appendix~\ref{app:diagnostics}).

\subsection{Clustered and Personalized FL}
\label{sec:exp-structure}
\paragraph{Group-structured heterogeneity.}
In these experiments we group clients in a way, that within a group the clients have IID data, and non-IID compared to clients from other groups. We consider two settings with and without ground-truth number of groups $J$ provided, our method supports both. \Cref{tab:clustering} compares accuracy and Adjusted Rand Index (ARI) on strict CIFAR-10 label skew, grouped FEMNIST, and Fed-ISIC2019. With the true group count, \crfg{} has the highest average accuracy of all methods across the five settings (73.4\%, versus 71.6\% for the strongest baseline, FLUX). It recovers the CIFAR-10 groups exactly (ARI 1.00), reaching 87.5\% and 95.6\% at medium and high severity, versus 76.9\% and 90.7\% for the strongest baselines given the same count. It also recovers FEMNIST label-permutation groups exactly, but fails on rotation groups (ARI 0.07, accuracy 48.5\%). With the true count, mixed evidence is the most accurate method under high-severity CIFAR-10 label skew (95.7\%) and on label-swapped FEMNIST (64.7\%), and improves on geometric evidence for rotated FEMNIST (60.3\%) even without accurate definition of those groups. Without the true count, \crfh{} is the only method that recovers the medium-severity CIFAR-10 groups (ARI 0.95) and the most accurate one there (84.8\%, versus 81.0\% for FLUX). On Fed-ISIC2019, \crfg{} finds two groups and reaches 72.4\%, on par with FLUX (72.6\%).

\begin{table}[t]
\centering
\caption{Clustered FL on group-structured heterogeneity: client test accuracy (\%) and adjusted Rand index (ARI) between discovered and true groups, 3 seeds. Methods in the oracle-$J$ groups receive the true number of clusters, the others estimate it.}
\label{tab:clustering}
\small
\setlength{\tabcolsep}{4pt}
\resizebox{\ifdim\width>\textwidth\textwidth\else\width\fi}{!}{%
\begin{tabular}{lcccccccccccc}
\toprule
Method & \multicolumn{2}{c}{CIFAR $P(y)$ strict, med.} & \multicolumn{2}{c}{CIFAR $P(y)$ strict, high} & \multicolumn{2}{c}{FEMNIST, rotated} & \multicolumn{2}{c}{FEMNIST, label-swapped} & \multicolumn{2}{c}{Fed-ISIC2019 (centres)} & \multicolumn{2}{c}{average} \\
\cmidrule(lr){2-3} \cmidrule(lr){4-5} \cmidrule(lr){6-7} \cmidrule(lr){8-9} \cmidrule(lr){10-11} \cmidrule(lr){12-13}
 & acc $\uparrow$ & ARI $\uparrow$ & acc $\uparrow$ & ARI $\uparrow$ & acc $\uparrow$ & ARI $\uparrow$ & acc $\uparrow$ & ARI $\uparrow$ & acc $\uparrow$ & ARI $\uparrow$ & acc $\uparrow$ & ARI $\uparrow$ \\
\midrule
\multicolumn{13}{l}{\emph{Baselines}} \\
FedAvg & 46.5{\scriptsize$\pm$4.8} & -- & 35.6{\scriptsize$\pm$19.4} & -- & 60.8{\scriptsize$\pm$3.6} & -- & 35.0{\scriptsize$\pm$1.3} & -- & 68.3{\scriptsize$\pm$0.7} & -- & 49.2 & -- \\
CFL & 39.8{\scriptsize$\pm$8.4} & 0.00 & 39.2{\scriptsize$\pm$16.8} & 0.00 & \textbf{64.9{\scriptsize$\pm$4.3}} & \underline{0.53} & 63.7{\scriptsize$\pm$10.4} & 0.87 & 68.3{\scriptsize$\pm$0.7} & 0.00 & 55.2 & 0.28 \\
FLUX & 81.0{\scriptsize$\pm$3.0} & 0.49 & 92.4{\scriptsize$\pm$5.4} & 0.47 & 49.9{\scriptsize$\pm$1.7} & 0.02 & 62.2{\scriptsize$\pm$2.0} & 0.38 & \textbf{72.6{\scriptsize$\pm$1.6}} & \textbf{0.83} & \underline{71.6} & 0.44 \\
\midrule
\multicolumn{13}{l}{\emph{Ours}} \\
\crfg{} & 38.1{\scriptsize$\pm$5.9} & 0.00 & 46.8{\scriptsize$\pm$10.7} & 0.25 & 55.9{\scriptsize$\pm$2.4} & 0.02 & 36.1{\scriptsize$\pm$2.7} & 0.00 & \underline{72.4{\scriptsize$\pm$1.3}} & \underline{0.58} & 49.9 & 0.17 \\
\crfv{} & 59.9{\scriptsize$\pm$33.6} & 0.44 & 63.9{\scriptsize$\pm$14.3} & 0.47 & 44.3{\scriptsize$\pm$27.8} & 0.04 & 27.7{\scriptsize$\pm$7.3} & 0.00 & 69.2{\scriptsize$\pm$0.8} & 0.00 & 53.0 & 0.19 \\
\crfh{} & \underline{84.8{\scriptsize$\pm$5.0}} & \underline{0.95} & 60.7{\scriptsize$\pm$18.4} & 0.24 & 57.8{\scriptsize$\pm$3.6} & 0.02 & 33.8{\scriptsize$\pm$3.3} & 0.00 & 71.7{\scriptsize$\pm$1.4} & \underline{0.58} & 61.8 & 0.36 \\
\midrule
\multicolumn{13}{l}{\emph{Baselines with oracle $J$}} \\
IFCA & 76.9{\scriptsize$\pm$6.2} & 0.80 & 77.1{\scriptsize$\pm$11.1} & 0.67 & 62.0{\scriptsize$\pm$4.2} & 0.33 & 58.5{\scriptsize$\pm$2.7} & 0.67 & 57.3{\scriptsize$\pm$1.5} & 0.00 & 66.4 & 0.49 \\
FedSoft & 74.2{\scriptsize$\pm$3.5} & 0.68 & 86.1{\scriptsize$\pm$12.9} & 0.60 & 60.9{\scriptsize$\pm$1.4} & \textbf{0.95} & 61.4{\scriptsize$\pm$3.8} & \textbf{1.00} & 65.9{\scriptsize$\pm$1.6} & 0.44 & 69.7 & \textbf{0.73} \\
FedRC & 72.3{\scriptsize$\pm$5.0} & 0.75 & 84.8{\scriptsize$\pm$7.8} & \underline{0.93} & \underline{62.3{\scriptsize$\pm$0.6}} & 0.00 & 40.9{\scriptsize$\pm$10.6} & 0.15 & 59.1{\scriptsize$\pm$1.6} & 0.22 & 63.9 & 0.41 \\
FedFomo & 63.4{\scriptsize$\pm$7.2} & -- & 88.4{\scriptsize$\pm$7.1} & -- & 41.7{\scriptsize$\pm$0.5} & -- & 45.6{\scriptsize$\pm$7.2} & -- & 69.4{\scriptsize$\pm$0.7} & -- & 61.7 & -- \\
FLUX & 75.7{\scriptsize$\pm$8.5} & 0.63 & 90.7{\scriptsize$\pm$10.0} & 0.88 & 56.5{\scriptsize$\pm$2.3} & 0.02 & 56.7{\scriptsize$\pm$4.5} & 0.50 & 70.1{\scriptsize$\pm$1.3} & 0.48 & 69.9 & 0.50 \\
\midrule
\multicolumn{13}{l}{\emph{Ours with oracle $J$}} \\
\crfg{} & \textbf{87.5{\scriptsize$\pm$2.2}} & \textbf{1.00} & \underline{95.6{\scriptsize$\pm$2.6}} & \textbf{1.00} & 48.5{\scriptsize$\pm$4.9} & 0.07 & \underline{64.1{\scriptsize$\pm$5.9}} & \textbf{1.00} & 71.3{\scriptsize$\pm$0.4} & 0.44 & \textbf{73.4} & \underline{0.70} \\
\crfv{} & 57.9{\scriptsize$\pm$32.5} & 0.62 & 94.7{\scriptsize$\pm$2.6} & \textbf{1.00} & 23.6{\scriptsize$\pm$16.3} & 0.02 & 53.7{\scriptsize$\pm$7.1} & \underline{0.89} & 64.9{\scriptsize$\pm$3.2} & 0.27 & 59.0 & 0.56 \\
\crfh{} & 58.6{\scriptsize$\pm$33.1} & 0.47 & \textbf{95.7{\scriptsize$\pm$3.0}} & \textbf{1.00} & 60.3{\scriptsize$\pm$2.4} & 0.11 & \textbf{64.7{\scriptsize$\pm$5.4}} & \textbf{1.00} & 71.1{\scriptsize$\pm$1.0} & 0.47 & 70.1 & 0.61 \\
\bottomrule
\end{tabular}}
\end{table}

\paragraph{Per-client heterogeneity.}
We consider no group-structure among clients. The goal is to train the best possible per-client models while still benefiting from collaboration across clients. FedCRF uses the affinities for client-specific mixing (\cref{tab:personal}). 
FedCRF with mixed evidence achieves the highest four-setting average (74.6\%), followed by functional evidence (74.1\%) and pFedGraph (72.7\%). The clearest gain occurs under class-conditional feature shift $P(x|y)$, where mixed evidence reaches 84.8\%, compared with 81.8\% for pFedGraph. Functional evidence is also the most accurate under label skew, reaching 93.6\%, slightly above local training (93.4\%) and FLUX (93.3\%).

\begin{table}[t]
\centering
\caption{Personalized FL on per-client heterogeneity (CIFAR-10 at ANDA high severity; FEMNIST with 20 writers): client test accuracy (\%, $\uparrow$), 3 seeds.}
\label{tab:personal}
\small
\setlength{\tabcolsep}{4pt}
\resizebox{\ifdim\width>\textwidth\textwidth\else\width\fi}{!}{%
\begin{tabular}{lccccc}
\toprule
Method & \multicolumn{3}{c}{ANDA high severity (per-client shift)} &   &   \\
\cmidrule(lr){2-4}
 & CIFAR $P(y)$ & CIFAR $P(y\mid x)$ & CIFAR $P(x\mid y)$ & FEMNIST & avg \\
\midrule
\multicolumn{6}{l}{\emph{Baselines}} \\
FedAvg & 14.2{\scriptsize$\pm$3.0} & 52.5{\scriptsize$\pm$2.3} & 55.4{\scriptsize$\pm$12.1} & 74.0{\scriptsize$\pm$2.9} & 49.0 \\
Local only & \underline{93.4{\scriptsize$\pm$2.7}} & 39.5{\scriptsize$\pm$2.9} & 74.9{\scriptsize$\pm$3.3} & 56.9{\scriptsize$\pm$2.4} & 66.2 \\
FedAvg + FT & 91.4{\scriptsize$\pm$2.1} & \textbf{58.4{\scriptsize$\pm$2.4}} & 74.3{\scriptsize$\pm$8.8} & 62.8{\scriptsize$\pm$3.9} & 71.7 \\
Ditto & 91.6{\scriptsize$\pm$1.8} & 45.7{\scriptsize$\pm$1.7} & 68.6{\scriptsize$\pm$22.9} & 67.7{\scriptsize$\pm$4.5} & 68.4 \\
pFedGraph & 86.3{\scriptsize$\pm$1.7} & 51.9{\scriptsize$\pm$3.4} & 81.8{\scriptsize$\pm$3.9} & 70.6{\scriptsize$\pm$2.7} & 72.7 \\
IFCA & 20.8{\scriptsize$\pm$0.6} & 53.0{\scriptsize$\pm$2.7} & 68.9{\scriptsize$\pm$10.7} & \underline{74.3{\scriptsize$\pm$3.2}} & 54.2 \\
CFL & 12.9{\scriptsize$\pm$1.4} & \underline{53.7{\scriptsize$\pm$1.2}} & 62.1{\scriptsize$\pm$8.6} & 72.1{\scriptsize$\pm$3.4} & 50.2 \\
FedSoft & 90.7{\scriptsize$\pm$2.6} & 41.9{\scriptsize$\pm$2.2} & 57.6{\scriptsize$\pm$16.9} & 61.2{\scriptsize$\pm$4.1} & 62.8 \\
FedRC & 18.4{\scriptsize$\pm$7.6} & 53.3{\scriptsize$\pm$1.7} & 79.3{\scriptsize$\pm$1.2} & \textbf{76.0{\scriptsize$\pm$2.7}} & 56.8 \\
FedFomo & 90.5{\scriptsize$\pm$2.5} & 22.4{\scriptsize$\pm$15.4} & 60.5{\scriptsize$\pm$4.9} & 48.7{\scriptsize$\pm$1.7} & 55.5 \\
FLUX & 93.3{\scriptsize$\pm$1.9} & 44.4{\scriptsize$\pm$2.9} & 75.2{\scriptsize$\pm$3.0} & 62.3{\scriptsize$\pm$3.4} & 68.8 \\
\midrule
\multicolumn{6}{l}{\emph{Ours}} \\
\crfg{} & 78.2{\scriptsize$\pm$9.3} & 51.0{\scriptsize$\pm$2.5} & \underline{83.6{\scriptsize$\pm$1.7}} & 66.4{\scriptsize$\pm$1.2} & 69.8 \\
\crfv{} & \textbf{93.6{\scriptsize$\pm$2.3}} & 51.9{\scriptsize$\pm$2.4} & 81.3{\scriptsize$\pm$1.8} & 69.5{\scriptsize$\pm$0.7} & \underline{74.1} \\
\crfh{} & 90.6{\scriptsize$\pm$1.2} & 53.5{\scriptsize$\pm$2.2} & \textbf{84.8{\scriptsize$\pm$1.4}} & 69.4{\scriptsize$\pm$0.2} & \textbf{74.6} \\
\bottomrule
\end{tabular}}
\end{table}

\section{Discussion and limitations}
\label{sec:limitations}
FedCRF provides a common inference framework for client reliability and relationships that can be reused across global aggregation, Byzantine robustness, clustered FL, and personalized FL. While on average FedCRF consistently outperforms baselines, some specialized approaches can edge it in specific cases.


While FedCRF is effective against the attacks considered on CIFAR-10, its robustness does not transfer uniformly across datasets. In particular, the FEMNIST results (see Appendix \cref{tab:robust-natural}) show that some backdoor attacks can retain high attack success despite strong clean accuracy, a failure mode also observed for many robust aggregation baselines. This suggests a broader difficulty in detecting localized or trigger-dependent behavior from coarse whole-model update statistics or clean proxy evaluation. 
A promising extension is therefore to enrich both sources of evidence: geometric evidence could be constructed at a finer granularity, for example layer-wise or over parameter groups, so that localized anomalies are less likely to be diluted in the overall model signal. Functional evidence could actively probe returned models using perturbed or augmented proxy inputs and measure their prediction stability or disagreement, potentially exposing behaviors that remain hidden on clean inputs. In our work we also use a fixed set of hyperparameters, which potentially can be made trainable for even better versatility.

Clustered and personalized cases depend on the affinity representation and group-count estimate, both of which may fail in some settings. 
The CRF on a dense graph introduces computational overhead over standard FedAvg. Appendix~\ref{app:cost} shows, however, that this overhead is comparable to several lightweight server-side baselines and lower than many specialized alternatives. 


\section{Conclusion}
\label{sec:conclusion}
FedCRF formulates server-side aggregation as structured probabilistic inference over client contributions, jointly modeling individual reliability and inter-client relationships from geometric or functional evidence. The resulting reliabilities and affinities provide a common representation that can be reused across global aggregation, Byzantine robustness, clustered FL, and personalized FL. Across heterogeneous client quality, partial participation, malicious updates, and structured or client-specific heterogeneity, FedCRF achieves competitive or leading performance against specialized methods. These findings support structured reliability modeling as a flexible foundation for federated aggregation across diverse FL settings. 

\subsection*{AI use statement}
We used large language model (LLM) assistants in three ways: to help implement the method and the baselines, to help set up, run, and report the experiments (experiment configurations, job orchestration, and the scripts that build the result tables), and to proofread and edit the text. The research questions, the method, the experimental design, and the interpretation of the results are our own. We take full responsibility for the content of the paper.

\subsection*{Reproducibility statement}
The code, the batch configuration of every experiment, and the result workbooks behind every table are available at \url{https://anonymous.4open.science/r/fedcrf-code-0044}. The repository contains one script per table and for the diagnostics figure, each of which reruns the corresponding experiments and regenerates the table, and a script that regenerates all tables and the figure from the released results without training.

\bibliography{references}
\bibliographystyle{iclr2027_conference}

\newpage
\appendix
\section*{Appendix}
\section{Related work}
\label{sec:related}
\paragraph{Server-side weighting.}
FedNova normalizes unequal local work \citep{wang2020fednova}, FedLAW learns aggregation weights and a global step size on proxy data \citep{li2023fedlaw}, and FedLWS applies layer-wise shrinkage from update statistics \citep{shi2025fedlws}. 
FedCRF instead treats aggregation weights as outputs of inference over latent client reliabilities, combining individual scores with pairwise compatibility.

\paragraph{Robust aggregation.}
Robust aggregation is one family within the broader landscape of FL defenses against adversarial behavior \citep{fenoglioSecurityPrivacyFLSurvey2026e}.
Coordinate-wise median, trimmed mean \citep{yin2018byzantine}, Krum and Multi-Krum \citep{blanchard2017machine}, and RFA \citep{pillutla2022robust} use robust statistics or update selection. Zeno scores updates using validation loss \citep{xie2019zeno}, while FLTrust compares them with an update from a server-held root dataset \citep{cao2021fltrust}. Our functional potentials similarly use server data and add predictive agreement between clients; on CIFAR-10, functional FedCRF matches the strongest server-data defense, Zeno, while also providing the client affinities used for clustering and personalization. Similarity also underlies FoolsGold's detection of coordinated clients \citep{fung2020foolsgold} and FLAME's update clustering \citep{nguyen2022flame}. FedCRF uses similarity to couple reliability labels, with a gate that limits reinforcement between individually unreliable clients.

\paragraph{Clustered and personalized FL.}
Clustered methods learn group models through client assignments or memberships (IFCA, FedSoft, FedRC \citep{ghosh2020ifca,ruan2022fedsoft,guo2024fedrc}), update-similarity partitions (CFL \citep{sattler2019cfl}), or data descriptors (FLUX \citep{fenoglioFLUX2025c}). Personalized methods regularize client models toward a shared model (Ditto, pFedMe \citep{li2021ditto,dinh2020pfedme}) or mix models using client-specific weights (FedFomo \citep{zhang2021fedfomo}). Graph-based approaches such as SFL and pFedGraph explicitly model client relationships \citep{chen2022sfl,ye2023pfedgraph}. FedCRF reuses the affinities constructed for reliability inference to support either clustering or personalized mixing, with inferred reliabilities adjusting client contributions.

\paragraph{CRFs and mean-field inference.}
Dense CRFs combine individual evidence with pairwise label compatibility through approximate inference \citep{krahenbuhl2011efficient}, which can also be unrolled as recurrent layers \citep{zheng2015crfasrnn}. FedCRF applies this formulation to clients, using latent reliability labels to determine their aggregation weights.

\section{Experimental setup}
\label{app:setup}

\subsection{Datasets and models}
\paragraph{CIFAR-10 and MNIST.}
CIFAR-10 contains 50{,}000 training and 10{,}000 test RGB images of size $32\times32$ in 10 classes. We use ResNet-9 and ANDA partitions (\cref{app:anda}). The ablations in \cref{app:structure,app:ablations} also use MNIST with the same partitioning scheme and an MLP with hidden layers of size 256 and 128.

\paragraph{FEMNIST.}
We use LEAF's Federated EMNIST \citep{caldas2018leaf}: $28\times28$ grayscale images in 62 classes, grouped by writer. We retain writers with at least 100 examples, reserve a random 20\% of writers for the official test set, and sample clients from the rest ($K=20$ by default). The LEAF CNN has two $5\times5$ convolutional layers with 32 and 64 channels, each followed by max pooling, and a 2048-unit fully connected layer before the classifier.

\paragraph{Grouped FEMNIST.}
For \cref{tab:clustering}, we randomly assign 20 writers to four groups of five. Each group applies either a rotation of $g\cdot90^\circ$ or a fixed permutation of the 62 labels, with group $g=0$ using the identity permutation. The transformation applies to all client splits; proxy and official test data remain unchanged. Group assignments provide the reference labels for ARI.

\paragraph{Fed-ISIC2019.}
This FLamby benchmark \citep{flamby2022} contains 8 skin-lesion classes from 6 acquisition centres. We use FLamby's train/test split and resizing, followed by a $200\times200$ centre crop, and an ImageNet-pretrained EfficientNet-B0 with a new classifier. Centres are divided into 18 sub-clients in proportion to their size, with at least one per centre. Centre identity defines the reference groups. We also report balanced accuracy because of class imbalance.

\paragraph{Splits, proxy set, and metrics.}
Client data is split 80/10/10 for training, validation, and testing. Early stopping uses sample-weighted client validation loss. Client test accuracy is sample-weighted across clients, each evaluated with its assigned global, cluster, or personalized model. FedCRF's functional variants, FedLAW, FLTrust, and Zeno share a proxy set of 10 examples per class from the official test split; these examples are excluded from official test evaluation.

\subsection{Heterogeneity with ANDA}
\label{app:anda}
ANDA \citep{
liFederatedLearningProfile2025b} varies one factor of $P(x,y)$ across clients at a time. We use static partitions; paired values below specify medium and high severity.
\begin{itemize}
    \item \emph{Label skew} $P(y)$: clients receive equal shares of the remaining data pool, sampled with class probabilities $\mathrm{softmax}(n_c^{\,s})$, where $n_c$ is the remaining class count and $s$ is sampled uniformly from $[0.5,1]$ / $[1,3]$ for each client.
    \item \emph{Feature skew} $P(x)$: client-specific distributions over rotations and colours, using 2 rotations and 2 colours / 4 rotations and 3 colours.
    \item \emph{Concept shift} $P(y\mid x)$: each client permutes labels within a randomly chosen pool of 3 / 5 classes.
    \item \emph{Class-conditional feature shift} $P(x\mid y)$: each client rotates and recolours 3 / 5 randomly chosen classes, using 2 rotations and 2 colours / 4 rotations and 3 colours.
\end{itemize}
Default partitions assign each client its own shift. The \emph{strict} variants instead use known groups: strict label skew assigns clients to three label subsets of 4 / 2 classes, while strict feature skew assigns a rotation--colour combination from 2 rotations and 2 colours / 4 rotations and 3 colours.

\subsection{Client quality and participation}
For \cref{tab:quality} and the diagnostics in \cref{app:diagnostics}, we corrupt only the training split of 30\% of randomly chosen clients. Label noise replaces 50\% or 80\% of labels with a uniformly sampled different class; input noise adds Gaussian noise with standard deviation 1 to normalized inputs. For \cref{tab:participation}, each round samples $pK$ clients uniformly. Sampling depends only on the seed and round, ensuring identical cohorts across methods.

\subsection{Attacks}
We select 20\% of clients as malicious once per run; they attack from the first round.
\begin{itemize}
    \item \emph{Sign flipping}: submit $w_t-2u_i$ instead of $w_t+u_i$.
    \item \emph{ALIE} \citep{baruch2019alie}: submit the coordinate-wise mean of honest updates minus $z$ times their standard deviation, with $z$ determined by the participating and malicious client counts.
    \item \emph{Targeted label flipping}: change class 1 to class 0. ASR is the fraction of class-1 test examples predicted as class 0.
    \item \emph{Model replacement} \citep{bagdasaryan2020backdoor}: use the same backdoor training and scale the submitted update by 10.
    \item \emph{Neurotoxin} \citep{zhang2022neurotoxin}: use the same backdoor training and restrict the update to coordinates outside the largest 5\% of the previous global update in absolute magnitude.
\end{itemize}

\subsection{Baselines}
\label{app:baselines}
All methods use the training setup in \cref{sec:setup}, with the following aggregation or local-objective changes.
\begin{itemize}
    \item \emph{Global aggregation.} FedAvg \citep{mcmahan2017communication} uses sample-count weights. FedProx \citep{li2020federated} adds $\frac{\mu}{2}\|w-w_t\|^2$ to the local objective ($\mu=0.01$). FedNova \citep{wang2020fednova} normalizes by local step count. FedLAW \citep{li2023fedlaw} learns weights and global shrinkage on the proxy set using 100 server epochs of Adam. FedLWS \citep{shi2025fedlws} applies layer-wise shrinkage ($\beta=0.03$). 
    \item \emph{Robust aggregation.} We use coordinate-wise median and trimmed mean \citep{yin2018byzantine}, trimming 10\% from each side. Krum and Multi-Krum \citep{blanchard2017machine} assume $f=\lfloor0.2|S_t|\rfloor$ attackers and select one or $|S_t|-f-2$ updates by nearest-neighbour distance scores. RFA \citep{pillutla2022robust} computes a sample-weighted geometric median with Weiszfeld iterations; the geometric-median baseline uses uniform weights. Fed-NGA averages unit-normalized updates.
    \item \emph{Robust aggregation with server data.} FLTrust \citep{cao2021fltrust} trains a root update $u_0$ using the median client step count, scores clients by $\mathrm{ReLU}(\cos(u_i,u_0))$, rescales their updates to $\|u_0\|$, and averages with these scores. Zeno \citep{xie2019zeno} scores proxy-loss decrease minus $\rho\|u_i\|^2$ ($\rho=5\cdot10^{-4}$) and averages the best $|S_t|-f$ updates.
    \item \emph{Clustered FL.} IFCA \citep{ghosh2020ifca} assigns clients by local loss; CFL \citep{sattler2019cfl} recursively partitions update similarities when training stalls; FedSoft \citep{ruan2022fedsoft} uses soft memberships; FedRC \citep{guo2024fedrc} models distribution shifts; and FLUX \citep{fenoglioFLUX2025c} clusters data descriptors. IFCA, FedSoft, FedRC, and oracle-count FLUX receive the true group count where defined, and two clusters otherwise. Adaptive FLUX estimates the count.
    \item \emph{Personalized FL.} Local training uses no communication. FedAvg with fine-tuning evaluates each client's locally updated global model. Ditto \citep{li2021ditto} regularizes a personal model toward the FedAvg model ($\lambda=0.1$). pFedGraph \citep{ye2023pfedgraph} uses update similarity to infer a collaboration graph (prior $\alpha=0.8$) and a cosine regularizer toward the graph-weighted model ($\lambda=0.01$). FedFomo \citep{zhang2021fedfomo} weights candidate models by local validation improvement, using the true group count as its candidate count where groups are defined.
\end{itemize}

\subsection{FedCRF settings}
\label{app:fedcrf-settings}
Unless varied in an ablation, we use the geometric coefficients $(2,1,0.5)$, $\lambda=0.5$, $\theta_G=0.5$, $\theta_F=\tau=1$, $\mathcal{Y}=\{0,0.25,1\}$, and $R=5$. Mixed evidence uses equal weights; functional evidence uses the proxy set above.

\emph{Clustering.} We run five global warm-up rounds with all clients and no reliability gate, averaging $-\log A_{ij}$ into distances $\bar d_{ij}$. Local scaling uses each client's seventh-nearest-neighbour distance \citep{zelnikmanor2004selftuning}. Spectral clustering uses the normalized Laplacian, with the group count selected by the largest eigengap among the first 12 eigenvalues on CIFAR-10 and grouped FEMNIST, or 8 on Fed-ISIC2019. Oracle-count variants use the true count. Each group then trains with FedCRF aggregation.

\emph{Personalization.} After the same warm-up, clients maintain individual models. Each round updates the ungated distances by an exponential moving average with rate 0.3, recomputes locally scaled affinities, and mixes models using \cref{eq:crfp}, with all clients as neighbours, $T=1$, and $s=10$.

\clearpage
\section{Algorithm}
\label{app:algorithm}
\Cref{alg:fedcrf} separates one round into local training, evidence construction, inference, and aggregation. \textsc{ClientUpdate} performs the configured local training; evidence formulas are given in \cref{sec:method-evidence}. All client distributions in iteration $r+1$ use the distributions from iteration $r$. In computation, reliability scores are clipped to $[\epsilon,1-\epsilon]$, scale denominators are bounded below by $\epsilon$, and label normalization uses a numerically stable softmax. Structure-discovery settings are in \cref{app:fedcrf-settings}.

\begin{algorithm}[h]
\caption{FedCRF: one communication round.}
\label{alg:fedcrf}
\begin{algorithmic}[1]
\REQUIRE Global model $w_t$; clients $S_t$ with sample counts $n_i$
\REQUIRE Evidence mode $m\in\{G,F,G{+}F\}$; proxy set $\Dval$ if $m\neq G$; parameters $\mathcal{Y},\lambda,\theta_G,\theta_F,\tau,R,\epsilon$
\ENSURE Global model $w_{t+1}$, client reliabilities $\bar y$, and affinity matrix $A$
\STATE \textbf{1. Train clients and collect updates}
\FORALL{$i\in S_t$ \textbf{in parallel}}
    \STATE $w_{t+1}^{i}\leftarrow\textsc{ClientUpdate}(i,w_t)$
\ENDFOR
\STATE $u_i\leftarrow w_{t+1}^{i}-w_t$, \quad $\alpha_i^{\mathrm{FA}}\leftarrow n_i/\sum_{j\in S_t}n_j$ for each client $i$
\STATE \textbf{2. Construct evidence on the server}
\IF{$m=G$}
    \STATE Compute $(\rho,A)=(\rho^G,A^G)$ from the updates using \cref{eq:rho-geo}
\ELSIF{$m=F$}
    \STATE Compute $(\rho,A)=(\rho^F,A^F)$ on $\Dval$ using \cref{eq:rho-functional,eq:affinity-val}
\ELSE
    \STATE Compute both $(\rho^G,A^G)$ and $(\rho^F,A^F)$ using the same formulas
    \STATE $\rho_i\leftarrow(\rho_i^G+\rho_i^F)/2$, \quad $A_{ij}\leftarrow\sqrt{A_{ij}^G A_{ij}^F}$ for $i\neq j$
\ENDIF
\STATE Set $A_{ii}\leftarrow0$ for each client $i$
\STATE \textbf{3. Infer client reliabilities}
\STATE Initialize $q_i^{(0)}(y)\leftarrow\pi_i(y)$ from \cref{eq:label-init} for every $i$ and $y\in\mathcal{Y}$
\FOR{$r=0$ \TO $R-1$}
    \FORALL{$i\in S_t$ \textbf{in parallel}}
        \STATE $C_i^{(r)}(y)\leftarrow\lambda\sum_{j\neq i}A_{ij}\sum_{\ell\in\mathcal{Y}}q_j^{(r)}(\ell)(y-\ell)^2$ for each $y\in\mathcal{Y}$
        \STATE $\tilde q_i(y)\leftarrow\pi_i(y)\exp\!\big(-C_i^{(r)}(y)\big)$ for each $y\in\mathcal{Y}$
        \STATE $q_i^{(r+1)}(y)\leftarrow\tilde q_i(y)/\sum_{z\in\mathcal{Y}}\tilde q_i(z)$ for each $y\in\mathcal{Y}$
    \ENDFOR
\ENDFOR
\STATE $\bar y_i\leftarrow\sum_{y\in\mathcal{Y}}y\,q_i^{(R)}(y)$ for each client $i$
\STATE \textbf{4. Normalize weights and aggregate}
\STATE $v_i\leftarrow\alpha_i^{\mathrm{FA}}\bar y_i$ for each client $i$, \quad $Z\leftarrow\sum_{j\in S_t}v_j$
\IF{$Z>\epsilon$}
    \STATE $\alpha_i\leftarrow v_i/Z$ for each client $i$
\ELSE
    \STATE $\alpha_i\leftarrow\alpha_i^{\mathrm{FA}}$ for each client $i$ \COMMENT{fall back to FedAvg}
\ENDIF
\STATE $w_{t+1}\leftarrow\sum_{i\in S_t}\alpha_i w_{t+1}^{i}$
\RETURN $w_{t+1}$, $\bar y$, $A$
\end{algorithmic}
\end{algorithm}

\section{Byzantine robustness on FEMNIST}
\label{app:robust-femnist}
On FEMNIST, all three FedCRF variants are among the most accurate defenses (\cref{tab:robust-natural}). Mixed and functional evidence average 71.1\% clean accuracy over the five attacks, within 0.5 points of Zeno (71.6\%), and geometric evidence 70.6\%; all three are ahead of every robust aggregator without server data (RFA 69.5\%, trimmed mean 67.6\%, geometric median 65.0\%, Multi-Krum 62.8\%). FedCRF also neutralizes targeted label flipping (ASR 0.0--1.0\%) while keeping 71.5--74.1\% clean accuracy. The backdoors that scale or mask the malicious update (model replacement and Neurotoxin) remain effective against every aggregation rule that keeps high clean accuracy on FEMNIST, including Zeno, RFA, and trimmed mean (ASR above 95\%); only Krum, Multi-Krum, and FLTrust reduce them, at a large cost in clean accuracy (10.0\%, 62.8\%, and 4.9\% on average).

\begin{table}[t]
\centering
\caption{Byzantine robustness on FEMNIST (20 writers, 20\% fixed malicious clients, 3 seeds): clean accuracy (\%) under five attacks and attack success rate (ASR, \%) for the targeted attacks.}
\label{tab:robust-natural}
\small
\setlength{\tabcolsep}{4pt}
\resizebox{\ifdim\width>\textwidth\textwidth\else\width\fi}{!}{%
\begin{tabular}{lcccccccccc}
\toprule
Method & \multicolumn{6}{c}{Clean accuracy under attack $\uparrow$} & \multicolumn{4}{c}{Attack success rate $\downarrow$} \\
\cmidrule(lr){2-7} \cmidrule(lr){8-11}
 & sign flip & ALIE & targeted LF & model repl. & Neurotoxin & avg & targeted LF & model repl. & Neurotoxin & avg \\
\midrule
\multicolumn{11}{l}{\emph{Robust aggregation}} \\
Coord.\ median & 22.9{\scriptsize$\pm$30.8} & 63.9{\scriptsize$\pm$3.6} & 71.7{\scriptsize$\pm$4.8} & 69.6{\scriptsize$\pm$4.9} & 70.6{\scriptsize$\pm$3.8} & 59.7 & \textbf{0.0{\scriptsize$\pm$0.0}} & 95.9{\scriptsize$\pm$4.0} & 95.3{\scriptsize$\pm$5.2} & 63.7 \\
Trimmed mean & 47.7{\scriptsize$\pm$8.9} & 67.0{\scriptsize$\pm$4.5} & \textbf{75.1{\scriptsize$\pm$3.7}} & \textbf{75.0{\scriptsize$\pm$3.7}} & 73.2{\scriptsize$\pm$4.5} & 67.6 & 2.5{\scriptsize$\pm$2.4} & 99.8{\scriptsize$\pm$0.2} & 99.7{\scriptsize$\pm$0.2} & 67.3 \\
Krum & 12.1{\scriptsize$\pm$7.8} & 12.9{\scriptsize$\pm$16.7} & 4.5{\scriptsize$\pm$1.7} & 14.4{\scriptsize$\pm$14.2} & 6.1{\scriptsize$\pm$0.6} & 10.0 & \textbf{0.0{\scriptsize$\pm$0.0}} & \underline{2.0{\scriptsize$\pm$3.5}} & \textbf{0.0{\scriptsize$\pm$0.0}} & \textbf{0.7} \\
Multi-Krum & 64.0{\scriptsize$\pm$4.3} & 56.5{\scriptsize$\pm$12.6} & 60.4{\scriptsize$\pm$6.7} & 67.0{\scriptsize$\pm$5.4} & 66.0{\scriptsize$\pm$3.5} & 62.8 & 1.6{\scriptsize$\pm$2.7} & \textbf{1.7{\scriptsize$\pm$0.0}} & \underline{13.8{\scriptsize$\pm$7.5}} & \underline{5.7} \\
RFA & 57.1{\scriptsize$\pm$5.0} & \textbf{70.1{\scriptsize$\pm$2.3}} & 73.5{\scriptsize$\pm$2.9} & 72.9{\scriptsize$\pm$2.9} & \textbf{74.0{\scriptsize$\pm$3.9}} & 69.5 & 1.6{\scriptsize$\pm$2.7} & 99.9{\scriptsize$\pm$0.1} & 99.9{\scriptsize$\pm$0.1} & 67.1 \\
Geom.\ median & 40.5{\scriptsize$\pm$29.8} & 68.3{\scriptsize$\pm$2.8} & 73.0{\scriptsize$\pm$3.0} & 72.2{\scriptsize$\pm$4.6} & 70.9{\scriptsize$\pm$6.6} & 65.0 & 3.5{\scriptsize$\pm$3.1} & 100.0{\scriptsize$\pm$0.0} & 100.0{\scriptsize$\pm$0.0} & 67.8 \\
Fed-NGA & 59.8{\scriptsize$\pm$5.6} & 58.2{\scriptsize$\pm$0.6} & 62.3{\scriptsize$\pm$5.8} & 63.5{\scriptsize$\pm$3.5} & 63.4{\scriptsize$\pm$2.7} & 61.5 & \textbf{0.0{\scriptsize$\pm$0.0}} & 99.5{\scriptsize$\pm$0.7} & 99.1{\scriptsize$\pm$1.5} & 66.2 \\
\midrule
\multicolumn{11}{l}{\emph{With server data}} \\
FLTrust & 3.7{\scriptsize$\pm$0.9} & 3.7{\scriptsize$\pm$1.0} & 4.1{\scriptsize$\pm$0.9} & 7.2{\scriptsize$\pm$0.8} & 5.6{\scriptsize$\pm$2.2} & 4.9 & 22.2{\scriptsize$\pm$38.5} & 3.9{\scriptsize$\pm$5.2} & 34.2{\scriptsize$\pm$57.0} & 20.1 \\
Zeno & \textbf{71.6{\scriptsize$\pm$3.2}} & \underline{68.7{\scriptsize$\pm$4.5}} & 73.2{\scriptsize$\pm$4.7} & 71.3{\scriptsize$\pm$5.1} & \underline{73.4{\scriptsize$\pm$2.7}} & \textbf{71.6} & 1.4{\scriptsize$\pm$2.5} & 98.6{\scriptsize$\pm$1.7} & 100.0{\scriptsize$\pm$0.0} & 66.7 \\
\midrule
\multicolumn{11}{l}{\emph{Ours}} \\
\crfg{} & 66.8{\scriptsize$\pm$3.4} & 67.7{\scriptsize$\pm$4.6} & 71.5{\scriptsize$\pm$3.7} & \underline{73.7{\scriptsize$\pm$3.2}} & 73.3{\scriptsize$\pm$3.5} & 70.6 & \textbf{0.0{\scriptsize$\pm$0.0}} & 99.6{\scriptsize$\pm$0.5} & 99.9{\scriptsize$\pm$0.1} & 66.5 \\
\crfv{} & \underline{69.9{\scriptsize$\pm$4.5}} & 68.6{\scriptsize$\pm$3.9} & 72.8{\scriptsize$\pm$3.1} & 72.5{\scriptsize$\pm$3.5} & 71.6{\scriptsize$\pm$4.4} & 71.1 & \textbf{0.0{\scriptsize$\pm$0.0}} & 99.6{\scriptsize$\pm$0.3} & 100.0{\scriptsize$\pm$0.0} & 66.5 \\
\crfh{} & 67.4{\scriptsize$\pm$1.5} & 68.0{\scriptsize$\pm$3.1} & \underline{74.1{\scriptsize$\pm$2.8}} & 73.2{\scriptsize$\pm$2.6} & 72.8{\scriptsize$\pm$2.8} & \underline{71.1} & \underline{1.0{\scriptsize$\pm$1.6}} & 99.7{\scriptsize$\pm$0.3} & 99.9{\scriptsize$\pm$0.1} & 66.9 \\
\bottomrule
\end{tabular}}
\end{table}

\section{Self weight of personalized FedCRF}
\label{app:structure}
Increasing self-weight improves personalized FedCRF under several shifts (\cref{tab:selfweight}). The default $s=10$ has the highest client validation accuracy averaged over the six cells for mixed evidence (83.8\%, against 80.9\% for $s=3$ and 77.3\% for $s=1$); only under concept shift do $s=1$ and $s=3$ validate slightly better. For mixed evidence on CIFAR-10, raising $s$ from 1 to 3 to 10 increases accuracy from 78.0\% to 81.5\% to 90.6\% under label skew and from 55.2\% to 78.6\% to 84.8\% under class-conditional shift. Concept-shift accuracy changes little (53.5--54.3\%). The other evidence types benefit in the same way: at $s=10$, geometric evidence reaches 78.2\% under CIFAR-10 label skew (57.0\% at $s=1$) and functional evidence 93.6\%, the best result in that column, and geometric evidence is the most accurate variant under MNIST concept shift (81.9\%). Restricting geometric evidence to five neighbours at $s=1$ helps MNIST concept shift (77.6\% versus 66.8\%) but gives little improvement on CIFAR-10 label skew (58.1\% versus 57.0\%). These results motivate the default $s=10$ with all neighbours.
\begin{table}[t]
\centering
\caption{Personalized FedCRF on the per-client cells (client test accuracy, \%, $\uparrow$): self weight $s$ (default $s{=}10$), neighbour count $N_{nbr}$, and evidence.}
\label{tab:selfweight}
\small
\setlength{\tabcolsep}{4pt}
\resizebox{\ifdim\width>\textwidth\textwidth\else\width\fi}{!}{%
\begin{tabular}{lcccccc}
\toprule
Method & \multicolumn{3}{c}{MNIST high} & \multicolumn{3}{c}{CIFAR-10 high} \\
\cmidrule(lr){2-4} \cmidrule(lr){5-7}
 & $P(y)$ & $P(y\mid x)$ & $P(x\mid y)$ & $P(y)$ & $P(y\mid x)$ & $P(x\mid y)$ \\
\midrule
\crfg{} ($s{=}1$) & 94.7{\scriptsize$\pm$5.6} & 66.8{\scriptsize$\pm$4.1} & 95.4{\scriptsize$\pm$0.9} & 57.0{\scriptsize$\pm$17.5} & 53.0{\scriptsize$\pm$4.8} & 73.9{\scriptsize$\pm$8.1} \\
\crfg{} ($s{=}3$) & 95.8{\scriptsize$\pm$3.0} & 70.9{\scriptsize$\pm$2.9} & 94.0{\scriptsize$\pm$2.2} & 70.4{\scriptsize$\pm$16.0} & 54.2{\scriptsize$\pm$3.0} & 83.1{\scriptsize$\pm$1.5} \\
\crfg{} ($s{=}10$) & 99.5{\scriptsize$\pm$0.6} & \textbf{81.9{\scriptsize$\pm$2.2}} & 94.2{\scriptsize$\pm$0.4} & 78.2{\scriptsize$\pm$9.3} & 51.0{\scriptsize$\pm$2.5} & \underline{83.6{\scriptsize$\pm$1.7}} \\
\crfg{} ($s{=}1$, $N_{nbr}{=}5$) & 96.5{\scriptsize$\pm$2.8} & 77.6{\scriptsize$\pm$2.3} & 96.1{\scriptsize$\pm$0.3} & 58.1{\scriptsize$\pm$10.7} & \textbf{55.3{\scriptsize$\pm$1.9}} & 75.8{\scriptsize$\pm$2.1} \\
\crfv{} ($s{=}10$) & \textbf{99.8{\scriptsize$\pm$0.1}} & 73.9{\scriptsize$\pm$3.4} & 96.1{\scriptsize$\pm$0.4} & \textbf{93.6{\scriptsize$\pm$2.3}} & 51.9{\scriptsize$\pm$2.4} & 81.3{\scriptsize$\pm$1.8} \\
\crfh{} ($s{=}1$) & 97.2{\scriptsize$\pm$2.0} & 65.7{\scriptsize$\pm$2.5} & \underline{96.4{\scriptsize$\pm$0.4}} & 78.0{\scriptsize$\pm$8.7} & \underline{54.3{\scriptsize$\pm$3.1}} & 55.2{\scriptsize$\pm$31.9} \\
\crfh{} ($s{=}3$) & 99.1{\scriptsize$\pm$0.4} & 67.7{\scriptsize$\pm$3.2} & 96.3{\scriptsize$\pm$0.3} & 81.5{\scriptsize$\pm$14.1} & 53.9{\scriptsize$\pm$3.0} & 78.6{\scriptsize$\pm$3.0} \\
\crfh{} ($s{=}10$) & \underline{99.7{\scriptsize$\pm$0.2}} & \underline{78.0{\scriptsize$\pm$3.0}} & \textbf{96.4{\scriptsize$\pm$0.3}} & \underline{90.6{\scriptsize$\pm$1.2}} & 53.5{\scriptsize$\pm$2.2} & \textbf{84.8{\scriptsize$\pm$1.4}} \\
\bottomrule
\end{tabular}}
\end{table}

\section{Additional ablations}
\label{app:ablations}
\paragraph{Components.}
\Cref{tab:e2} compares full \crfg{} with direct reliability weighting ($\alpha_i\propto\alpha_i^{\mathrm{FA}}\rho_i$), removal of the reliability gate, and uniform base weights. In these benign settings no client needs to be down-weighted, so all variants perform comparably; the full model is the most accurate under CIFAR-10 $P(y)$ high (18.8\%). The structured components matter once some clients are unreliable: \cref{app:diagnostics} shows that under ALIE the pairwise term lowers the attackers' aggregation weight in every seed and improves the worst-case accuracy (\cref{tab:pairwise-alie}).

\paragraph{Inference.}
FedCRF is robust to its inference settings (\cref{tab:e3}). With $R=1$ or $R=10$ mean-field iterations instead of the default $R=5$, or with two or five labels instead of three, all variants lie within 0.8 points of one another at medium severity, and at high severity on CIFAR-10 the default is the most accurate (18.8\% against 10.4--13.7\%).

The Gaussian relaxation uses continuous labels, $y_i\in\mathbb{R}$, a quadratic unary potential $\psi_u(y_i,x_i)=(y_i-\rho_i)^2/(2\sigma_u^2)$, and the same pairwise potential. Its posterior mean and MAP estimate coincide:
\begin{equation}
    \bar{\boldsymbol{y}}=\big(I+2\lambda\sigma_u^2 L\big)^{-1}\boldsymbol{\rho},
    \qquad
    L=\mathrm{diag}(A\mathbf{1})-A.
\end{equation}
Here, $L$ is the affinity-graph Laplacian; we use $\sigma_u=1$ and clip the result to $[0,1]$. The relaxation smooths reliability scores over the graph without iterative label inference. The two formulations are comparable at medium severity, while discrete labels are clearly more accurate under high MNIST label skew: 85.7--86.5\% versus 76.8\% for geometric evidence, and 85.9\% versus 72.6\% for mixed evidence. This motivates the discrete label set used throughout.

\paragraph{Hyperparameters.}
On CIFAR-10 $P(y)$ medium, the 21 settings of $(\lambda,\theta_G)$ for \crfg{} and $(\theta_F,\tau)$ for \crfv{} span 84.0--85.6\% accuracy (\cref{tab:e4}) and 84.8--86.4\% mean client validation accuracy. This shows that sensitivity to a hyperparameter choice is modest for FedCRF.

\paragraph{Proxy set.}
FedCRF needs only a very small proxy set: on MNIST and CIFAR-10 $P(y)$ medium, one example per class performs comparably to ten, and fifty gives no further gain (\cref{tab:e7}). It also tolerates imperfect proxy data: label noise, class imbalance, and input noise in the proxy set reduce accuracy by at most 0.9 points relative to a clean proxy set.
\begin{table}[t]
\centering
\caption{Component ablation of \crfg{} in benign settings (client test accuracy, \%, $\uparrow$). Direct $\rho_i$ weighting sets $\alpha_i\propto\alpha_i^{\mathrm{FA}}\rho_i$, without the CRF labels or the pairwise term.}
\label{tab:e2}
\small
\setlength{\tabcolsep}{4pt}
\resizebox{\ifdim\width>\textwidth\textwidth\else\width\fi}{!}{%
\begin{tabular}{lcccc}
\toprule
Method & \multicolumn{2}{c}{CIFAR-10 $P(y)$} & \multicolumn{2}{c}{CIFAR-10 $P(x)$} \\
\cmidrule(lr){2-3} \cmidrule(lr){4-5}
 & med & high & med & high \\
\midrule
FedAvg & 84.5 & \underline{14.5} & \textbf{73.3} & \underline{67.2} \\
direct $\rho_i$ weighting & 84.6{\scriptsize$\pm$0.6} & 11.8{\scriptsize$\pm$4.6} & 72.6{\scriptsize$\pm$1.9} & 65.6{\scriptsize$\pm$1.3} \\
\crfg{} without gate & \textbf{85.0{\scriptsize$\pm$0.7}} & 12.0{\scriptsize$\pm$1.2} & 72.2{\scriptsize$\pm$0.5} & 67.1{\scriptsize$\pm$0.7} \\
\crfg{} uniform base weights & 84.5{\scriptsize$\pm$1.5} & 13.5{\scriptsize$\pm$4.4} & 72.4{\scriptsize$\pm$1.6} & \textbf{67.3{\scriptsize$\pm$1.5}} \\
\crfg{} & \underline{84.6{\scriptsize$\pm$0.9}} & \textbf{18.8{\scriptsize$\pm$6.8}} & \underline{72.9{\scriptsize$\pm$1.1}} & 67.2{\scriptsize$\pm$1.7} \\
\bottomrule
\end{tabular}}
\end{table}

\begin{table}[t]
\centering
\caption{Inference variants (client test accuracy, \%, $\uparrow$): mean-field iterations $R$, label-set size, and the Gaussian relaxation.}
\label{tab:e3}
\small
\setlength{\tabcolsep}{4pt}
\resizebox{\ifdim\width>\textwidth\textwidth\else\width\fi}{!}{%
\begin{tabular}{lcccc}
\toprule
Method & \multicolumn{2}{c}{MNIST $P(y)$} & \multicolumn{2}{c}{CIFAR-10 $P(y)$} \\
\cmidrule(lr){2-3} \cmidrule(lr){4-5}
 & med. & high & med. & high \\
\midrule
\crfg{} (default: $|\mathcal{Y}|=3$, $R=5$) & \underline{97.9{\scriptsize$\pm$0.0}} & 85.7{\scriptsize$\pm$4.5} & 84.6{\scriptsize$\pm$0.9} & \textbf{18.8{\scriptsize$\pm$6.8}} \\
mean field, $R=1$ & \textbf{98.0{\scriptsize$\pm$0.1}} & \textbf{86.5{\scriptsize$\pm$4.9}} & 84.6{\scriptsize$\pm$0.6} & 13.7{\scriptsize$\pm$3.5} \\
mean field, $R=10$ & 97.9{\scriptsize$\pm$0.1} & \underline{86.1{\scriptsize$\pm$5.2}} & \underline{85.0{\scriptsize$\pm$0.2}} & 10.8{\scriptsize$\pm$3.0} \\
mean field, $\mathcal{Y}=\{0,1\}$ & 97.8{\scriptsize$\pm$0.1} & 82.2{\scriptsize$\pm$4.4} & 84.6{\scriptsize$\pm$0.4} & 10.4{\scriptsize$\pm$1.0} \\
mean field, $|\mathcal{Y}|=5$ & 97.9{\scriptsize$\pm$0.1} & 82.0{\scriptsize$\pm$6.0} & \textbf{85.4{\scriptsize$\pm$0.7}} & 11.0{\scriptsize$\pm$3.7} \\
\crfg{} (Gaussian) & 97.8{\scriptsize$\pm$0.1} & 76.8{\scriptsize$\pm$6.7} & 84.9{\scriptsize$\pm$0.4} & 11.0{\scriptsize$\pm$0.3} \\
\crfh{} (default) & 97.9{\scriptsize$\pm$0.1} & 85.9{\scriptsize$\pm$3.6} & 84.7{\scriptsize$\pm$0.5} & \underline{17.5{\scriptsize$\pm$5.9}} \\
\crfh{} (Gaussian) & 97.8{\scriptsize$\pm$0.1} & 72.6{\scriptsize$\pm$5.3} & 84.7{\scriptsize$\pm$0.8} & 12.3{\scriptsize$\pm$1.0} \\
\bottomrule
\end{tabular}}
\end{table}

\begin{table}[t]
\centering
\caption{Hyperparameter sensitivity on CIFAR-10 $P(y)$ medium (client test accuracy, \%).}
\label{tab:e4}
\footnotesize
\setlength{\tabcolsep}{4pt}
\resizebox{\ifdim\width>\textwidth\textwidth\else\width\fi}{!}{%
\begin{tabular}{lc}
\toprule
 & accuracy $\uparrow$ \\
\midrule
\crfg{}: $\lambda=0.1$, $\theta_G=0.25$ & 84.5{\scriptsize$\pm$1.1} \\
\crfg{}: $\lambda=0.1$, $\theta_G=0.5$ & 84.4{\scriptsize$\pm$0.7} \\
\crfg{}: $\lambda=0.1$, $\theta_G=1$ & 84.2{\scriptsize$\pm$1.6} \\
\crfg{}: $\lambda=0.5$, $\theta_G=0.25$ & 84.9{\scriptsize$\pm$0.7} \\
\crfg{}: $\lambda=0.5$, $\theta_G=0.5$ & 84.6{\scriptsize$\pm$0.9} \\
\crfg{}: $\lambda=0.5$, $\theta_G=1$ & 85.6{\scriptsize$\pm$0.5} \\
\crfg{}: $\lambda=1$, $\theta_G=0.25$ & 84.2{\scriptsize$\pm$0.2} \\
\crfg{}: $\lambda=1$, $\theta_G=0.5$ & 85.0{\scriptsize$\pm$0.1} \\
\crfg{}: $\lambda=1$, $\theta_G=1$ & 84.7{\scriptsize$\pm$0.9} \\
\crfg{}: $\lambda=5$, $\theta_G=0.25$ & 84.3{\scriptsize$\pm$0.7} \\
\crfg{}: $\lambda=5$, $\theta_G=0.5$ & 84.5{\scriptsize$\pm$0.9} \\
\crfg{}: $\lambda=5$, $\theta_G=1$ & 84.8{\scriptsize$\pm$0.4} \\
\crfv{}: $\theta_F=0.5$, $\tau=0.5$ & 84.1{\scriptsize$\pm$0.6} \\
\crfv{}: $\theta_F=0.5$, $\tau=1$ & 84.8{\scriptsize$\pm$0.6} \\
\crfv{}: $\theta_F=0.5$, $\tau=2$ & 84.3{\scriptsize$\pm$0.3} \\
\crfv{}: $\theta_F=1$, $\tau=0.5$ & 84.2{\scriptsize$\pm$1.3} \\
\crfv{}: $\theta_F=1$, $\tau=1$ & 84.3{\scriptsize$\pm$1.2} \\
\crfv{}: $\theta_F=1$, $\tau=2$ & 84.6{\scriptsize$\pm$1.1} \\
\crfv{}: $\theta_F=2$, $\tau=0.5$ & 84.2{\scriptsize$\pm$0.4} \\
\crfv{}: $\theta_F=2$, $\tau=1$ & 84.0{\scriptsize$\pm$0.5} \\
\crfv{}: $\theta_F=2$, $\tau=2$ & 84.5{\scriptsize$\pm$0.6} \\
\bottomrule
\end{tabular}}
\end{table}

\begin{table}[t]
\centering
\caption{Proxy-set realism ($P(y)$ medium, client test accuracy, \%, $\uparrow$): proxy budget per class and proxy corruptions.}
\label{tab:e7}
\small
\setlength{\tabcolsep}{4pt}
\resizebox{\ifdim\width>\textwidth\textwidth\else\width\fi}{!}{%
\begin{tabular}{lcccc}
\toprule
Method & \multicolumn{2}{c}{MNIST} & \multicolumn{2}{c}{CIFAR-10} \\
\cmidrule(lr){2-3} \cmidrule(lr){4-5}
 & \crfv{} & \crfh{} & \crfv{} & \crfh{} \\
\midrule
reference (10/class, clean proxy) & 97.8{\scriptsize$\pm$0.0} & 97.9{\scriptsize$\pm$0.1} & 84.3{\scriptsize$\pm$1.2} & 84.7{\scriptsize$\pm$0.5} \\
proxy budget 1/class & 97.8{\scriptsize$\pm$0.0} & 97.9{\scriptsize$\pm$0.1} & 84.7{\scriptsize$\pm$1.0} & 84.6{\scriptsize$\pm$0.4} \\
proxy budget 5/class & 98.0{\scriptsize$\pm$0.0} & 98.0{\scriptsize$\pm$0.0} & 84.3{\scriptsize$\pm$0.2} & 84.5{\scriptsize$\pm$0.5} \\
proxy budget 50/class & 97.9{\scriptsize$\pm$0.1} & 97.9{\scriptsize$\pm$0.1} & 83.5{\scriptsize$\pm$0.3} & 85.0{\scriptsize$\pm$0.5} \\
label noise 20\% & 97.9{\scriptsize$\pm$0.1} & 98.0{\scriptsize$\pm$0.1} & 83.5{\scriptsize$\pm$0.7} & 84.5{\scriptsize$\pm$1.1} \\
label noise 50\% & 98.0{\scriptsize$\pm$0.0} & 98.0{\scriptsize$\pm$0.1} & 84.3{\scriptsize$\pm$0.5} & 84.2{\scriptsize$\pm$1.0} \\
class imbalance 20\% & 97.8{\scriptsize$\pm$0.1} & 97.9{\scriptsize$\pm$0.1} & 84.4{\scriptsize$\pm$1.1} & 84.8{\scriptsize$\pm$0.8} \\
class imbalance 50\% & 97.9{\scriptsize$\pm$0.1} & 98.0{\scriptsize$\pm$0.0} & 84.9{\scriptsize$\pm$0.9} & 83.9{\scriptsize$\pm$1.1} \\
input noise 0.2 & 97.9{\scriptsize$\pm$0.1} & 97.9{\scriptsize$\pm$0.1} & 84.8{\scriptsize$\pm$0.5} & 84.2{\scriptsize$\pm$0.4} \\
input noise 0.5 & 97.9{\scriptsize$\pm$0.1} & 97.9{\scriptsize$\pm$0.0} & 84.7{\scriptsize$\pm$0.3} & 83.8{\scriptsize$\pm$0.8} \\
\bottomrule
\end{tabular}}
\end{table}

\clearpage
\section{Computational cost}
\label{app:cost}
\paragraph{Complexity.}
FedCRF changes only the server step. With $N$ participating clients and $d$ model parameters, geometric evidence needs the coordinate-wise median of the updates ($O(Nd\log N)$), each update's distance to it ($O(Nd)$), and the pairwise cosine similarities ($O(N^2d)$). Functional evidence evaluates every returned model on the proxy set ($N$ forward passes over $10C$ examples for $C$ classes) and compares the predictions of every pair of clients ($O(N^2C^2)$). Mean-field inference on the fully connected graph costs $O(RN^2|\mathcal{Y}|^2)$, which is negligible. Like any rule that aggregates full updates, the server keeps the $N$ updates in memory. Clustering adds one eigendecomposition of an $N\times N$ matrix after the warm-up and then aggregates within each group, while personalization forms one mixture of the $N$ returned models per client ($O(N^2d)$ per round).

\paragraph{Measurements.}
We ran every method for four rounds without early stopping on one NVIDIA RTX 4090 GPU (one seed, no attack, the same proxy set for all methods) in the settings of the paper: CIFAR-10 with ResNet-9 (6.6M parameters) and FEMNIST with the LEAF CNN (6.6M parameters), each with 20 and 100 clients, and Fed-ISIC2019 with EfficientNet-B0 (4.0M parameters) and 18 clients. All clients participate in every round. \Cref{tab:overhead} reports the median wall-clock time of rounds 2--4, which includes local training, the server step, and evaluation, relative to FedAvg. 

\paragraph{Results.}
The server step of FedCRF is lightweight. Global FedCRF adds 0.7--0.9\,s to the server step with 20 clients on CIFAR-10 and FEMNIST (0.5--1.5\,s on Fed-ISIC2019, where functional evidence evaluates EfficientNet-B0 on the proxy set) and 3.1--5.3\,s with 100 clients, against 0.1\,s and 0.4--0.8\,s for FedAvg's own server step. How much this matters depends on the cost of local training. Per round, it is +10--13\% for ResNet-9 on CIFAR-10 with 20 clients, +29--40\% with 100 clients, and +1--4\% for EfficientNet-B0 on Fed-ISIC2019, but +95--148\% on FEMNIST, where each writer holds about 200 examples and a FedAvg round takes under a second with 20 clients. Geometric evidence is the cheapest variant, and functional evidence adds the proxy evaluations. With partial participation the server step handles only the sampled clients: with 100 CIFAR-10 clients at 10\% participation, global FedCRF adds 29--39\% to a 1.2\,s round, about as much as FedLWS (+28\%) and Zeno (+34\%) and less than trimmed mean (+45\%), RFA (+138\%), and FedLAW (+149\%).

Compared with the other server-side methods, global FedCRF costs about as much as FedLWS, FLTrust, and Zeno and less than the robust aggregators based on sorting or iterative estimation: coordinate-wise median and trimmed mean (+24--25\% on CIFAR-10 with 20 clients, +84--85\% with 100, and +176--344\% on FEMNIST), Multi-Krum (+22\%, +62\%, and +160--193\%), and the geometric median and RFA (+38\%, +103--104\%, and +394--549\%). FedLAW, which optimizes its weights on the proxy set, adds 42--333\%. 

Clustered FedCRF, which trains one model per group after the warm-up, adds 10\% on CIFAR-10 with 20 clients, 28\% with 100, and 84--90\% on FEMNIST, less than IFCA (+44--48\% and +101--102\%), FedSoft (+103--107\% and +142--144\%), and FedRC (+146\% and +178--184\%), which evaluate or train several models on every client, and comparable to CFL and FLUX. Personalized FedCRF adds 27\% and 91\% on CIFAR-10 with 20 and 100 clients, less than Ditto (+159\% and +172\%), FedFomo (+112\% and +316\%), and, with 20 clients, pFedGraph (+42\%; +98\% with 100), and 254--288\% on FEMNIST, more than Ditto and pFedGraph (+184--230\%) but far less than FedFomo ($\times$12--14). With the heavier EfficientNet-B0 on Fed-ISIC2019, all FedCRF variants add at most 6\% per round.

\begin{table}[t]
\centering
\caption{Computational cost relative to FedAvg ($\downarrow$): median wall-clock time of a training round (local training, server step, and evaluation) on one RTX 4090, rounds 2--4 of one seed without attack. The first row is FedAvg's time per round; the others give the increase over it, and values above +1000\% are given as a factor. All clients participate in every round. 
}
\label{tab:overhead}
\small
\setlength{\tabcolsep}{4pt}
\resizebox{\ifdim\width>\textwidth\textwidth\else\width\fi}{!}{%
\begin{tabular}{lccccc}
\toprule
Method & CIFAR-10, $K{=}20$ & CIFAR-10, $K{=}100$ & FEMNIST, $K{=}20$ & FEMNIST, $K{=}100$ & Fed-ISIC2019, $K{=}18$ \\
\midrule
FedAvg (time per round) & 7.35\,s & 10.20\,s & 0.75\,s & 3.56\,s & 38.95\,s \\
\midrule
\multicolumn{6}{l}{\emph{FedCRF}} \\
\crfg{} & +10\% & +29\% & +104\% & +95\% & +1\% \\
\crfv{} & +12\% & +37\% & +121\% & +144\% & +3\% \\
\crfh{} & +13\% & +40\% & +126\% & +148\% & +4\% \\
\crfg{}, clustered & +10\% & +28\% & +90\% & +84\% & +1\% \\
\crfh{}, personalized ($s{=}10$) & +27\% & +91\% & +254\% & +288\% & +6\% \\
\midrule
\multicolumn{6}{l}{\emph{Global aggregation}} \\
FedProx & +49\% & +38\% & +19\% & +16\% & +15\% \\
FedNova & +5\% & +11\% & +53\% & +48\% & +0\% \\
FedLWS & +10\% & +28\% & +94\% & +73\% & +1\% \\
FedLAW & +42\% & +129\% & +333\% & +242\% & +70\% \\
\midrule
\multicolumn{6}{l}{\emph{Robust aggregation}} \\
Coord.\ median & +25\% & +85\% & +179\% & +344\% & +2\% \\
Trimmed mean & +24\% & +84\% & +176\% & +342\% & +2\% \\
Krum & +20\% & +59\% & +177\% & +146\% & +2\% \\
Multi-Krum & +22\% & +62\% & +193\% & +160\% & +2\% \\
RFA & +38\% & +103\% & +549\% & +542\% & +10\% \\
Geom.\ median & +38\% & +104\% & +492\% & +394\% & +9\% \\
Fed-NGA & +23\% & +73\% & +217\% & +188\% & +2\% \\
FLTrust & +18\% & +36\% & +91\% & +81\% & +7\% \\
Zeno & +11\% & +35\% & +105\% & +95\% & +3\% \\
\midrule
\multicolumn{6}{l}{\emph{Clustered FL}} \\
IFCA & +44\% & +48\% & +101\% & +102\% & +146\% \\
CFL & +15\% & +40\% & +137\% & +126\% & +2\% \\
FedSoft & +107\% & +103\% & +144\% & +142\% & +162\% \\
FedRC & +146\% & +146\% & +178\% & +184\% & +588\% \\
FLUX & +9\% & +17\% & +78\% & +46\% & +7\% \\
\midrule
\multicolumn{6}{l}{\emph{Personalized FL}} \\
Local only & +5\% & +13\% & +19\% & +18\% & +1\% \\
FedAvg + FT & +4\% & +9\% & +30\% & +31\% & +0\% \\
Ditto & +159\% & +172\% & +184\% & +187\% & +116\% \\
pFedGraph & +42\% & +98\% & +230\% & +188\% & +6\% \\
FedFomo & +112\% & +316\% & $\times$12 & $\times$14 & +29\% \\
\bottomrule
\end{tabular}}
\end{table}

\section{Reliability diagnostics}
\label{app:diagnostics}
\begin{figure}[t]
    \centering
    \includegraphics[width=1\textwidth]{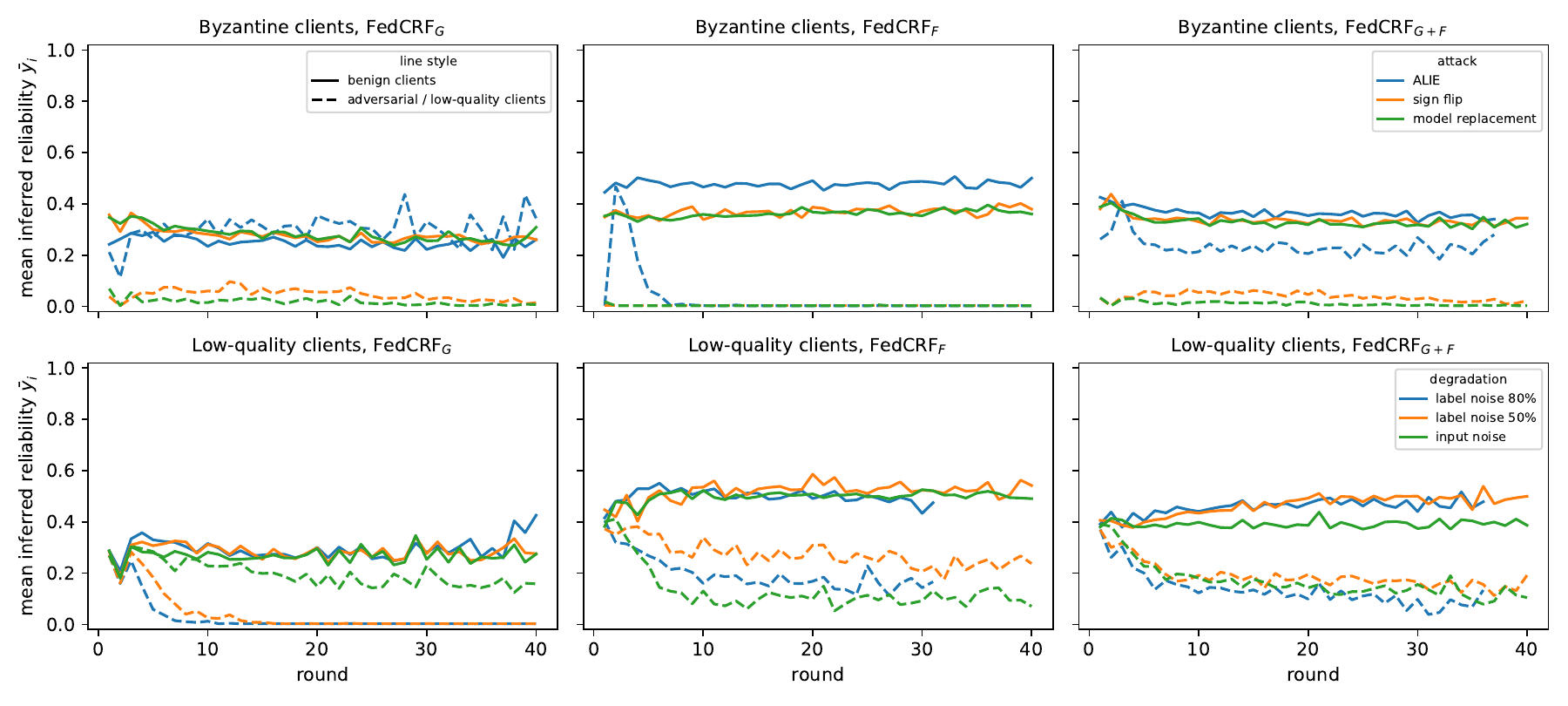}
    \caption{Mean inferred reliability on CIFAR-10 over 3 seeds. Dashed lines denote malicious clients (top, 20\%) or clients with corrupted training data (bottom, 30\%); solid lines denote the remaining clients. Columns compare geometric, functional, and mixed evidence.}
    \label{fig:diagnostics}
\end{figure}

\Cref{fig:diagnostics} tracks mean inferred reliability separately for attackers or corrupted clients and the remaining clients on CIFAR-10.

Geometric evidence assigns near-zero reliability to sign-flipping and model-replacement attackers from the first round. ALIE is designed to evade such statistics, since its updates stay within the spread of the benign ones; there, the pairwise potentials strengthen geometric evidence (\cref{tab:pairwise-alie}), and functional evidence suppresses all three attacks: ALIE attackers separate after two rounds and reach near-zero reliability by round seven, while benign clients settle at 0.4--0.5.

For corrupted data, geometric evidence drives the reliability of label-noisy clients to near zero, by round eight at 80\% label noise and by round fifteen at 50\%, and lowers input-noisy clients to 0.15--0.2 against about 0.3 for clean clients. Functional evidence keeps label-noisy clients at 0.15--0.25, since their updates still improve the proxy loss early in training; the two evidence sources are therefore complementary.

Mixed evidence combines both sources and lowers a client's reliability to around 0.15--0.2 when either source flags it. Inferred reliability also tracks each client's actual contribution: it correlates with per-client proxy-loss improvement on CIFAR-10 and high-severity MNIST (Pearson $r=0.44$--$0.74$), while on IID and medium-severity MNIST, where clients differ little in quality, there is little variation to track ($r\approx0$).

\paragraph{Pairwise potentials under ALIE.}
ALIE is the attack on which unary geometric evidence alone is least informative, which makes it a direct test of the importance of pairwise potentials. \Cref{tab:pairwise-alie} runs \crfg{} on CIFAR-10 under ALIE with ten seeds and three coupling strengths: $\lambda=0$ (unary potentials only), and $\lambda=\{0.5,2\}$. The pairwise term consistently shifts weight from the attackers to the benign clients: the attackers' share of the aggregate falls from 0.246 without coupling to 0.231 at $\lambda=0.5$ and to 0.184 at $\lambda=2$, and it is lower in every one of the ten seeds (the per-seed ranges 0.242--0.250, 0.223--0.240, and 0.181--0.186 do not overlap). Lower attacker weight translates into more accurate and more stable training: with $\lambda=2$, mean accuracy rises by 2.0 points (from 73.8\% to 75.8\%), the standard deviation across seeds falls from 4.1 to 1.8 points, and the worst seed improves from 63.4\% to 73.0\%. The default $\lambda=0.5$ already raises the worst seed to 69.3\% and lowers the standard deviation to 3.0 points. Coupling clients through their affinities thus lets the benign majority support one another's reliability, which a per-client score alone cannot do.

\begin{table}[h]
\centering
\caption{Pairwise coupling under ALIE: \crfg{} on CIFAR-10 ($P(y)$ medium, 20\% malicious clients), 10 seeds. Clean accuracy (\%, mean $\pm$ standard deviation and worst seed) and the attackers' share of the aggregation weight, averaged over rounds.}
\label{tab:pairwise-alie}
\small
\begin{tabular}{lccc}
\toprule
Coupling & accuracy $\uparrow$ & worst seed $\uparrow$ & attacker weight $\downarrow$ \\
\midrule
$\lambda=0$ (unary potentials only) & 73.8{\scriptsize$\pm$4.1} & 63.4 & 0.246 \\
$\lambda=0.5$  & 73.7{\scriptsize$\pm$3.0} & 69.3 & 0.231 \\
$\lambda=2$ & 75.8{\scriptsize$\pm$1.8} & 73.0 & 0.184 \\
\bottomrule
\end{tabular}
\end{table}

\end{document}